\documentclass[letterpaper, 10 pt, conference]{ieeeconf}

\makeatletter
\let\NAT@parse\undefined
\makeatother

\usepackage{hyperref}
\usepackage[T1]{fontenc}
\usepackage[utf8]{inputenc}
\usepackage{tabularx,ragged2e,booktabs,caption}
\newcolumntype{C}[1]{>{\Centering}m{#1}}

\usepackage{graphicx} 
\graphicspath{{figures/}}
\usepackage{epsfig} 
\usepackage{ragged2e}
\usepackage{amsmath}
\usepackage{amssymb}
\usepackage{epstopdf}
\usepackage{cite}
\usepackage[noend,ruled,linesnumbered]{algorithm2e}
\SetKwComment{Comment}{$\triangleright$\ } {}
\usepackage{multirow}
\usepackage{rotating}
\usepackage{subfigure}
\usepackage{color}
\usepackage{mysymbol}
\usepackage{url}
\usepackage{ltlfonts}	
\usepackage{ulem}
\usepackage{enumerate}
\usepackage{stmaryrd}
\usepackage{array}
\usepackage[numbers,sort&compress]{natbib}
\usepackage{mathtools} 
\usepackage{tikz}
\usepackage{ifthen}

\newboolean{arXiv}
\allowdisplaybreaks

\makeatletter
\renewcommand*{\@opargbegintheorem}[3]{\trivlist
  \item[\hskip \labelsep{\it\quad  #1\ #2:}] {\it(#3)}\ }
\makeatother
\newtheorem{theorem}{Theorem}[section]

\newtheorem{problem}{Problem}

\newtheorem{asmp}[theorem]{Assumption}
\newtheorem{exmp}{Example}

\newtheorem{defn}[theorem]{Definition}
\newtheorem{rem}[theorem]{Remark}

\newcommand{\level}[2]{\phi({#2}, {#1})}

\newcommand{\autop}{\ccalA_{\phi}}

\newcommand{\apsm}[3]{\pi_{\text{#1}}^{#2, #3}}

\newcommand{\tasks}[2]{\llbracket #1 \rrbracket_#2}
\newcommand{\tasksall}{\llbracket \phi \rrbracket}
\newcommand{\tasksallp}{\llbracket \phi' \rrbracket}
\newcommand{\seqans}[2]{\ccalS(\level{#1}{#2})}
\newcommand{\tphi}{\phi}
\newcommand{\tdot}[1]{\tikz\fill[#1] (0,0) circle (2pt);}
\newcommand{\modf}[1]{\boxed{#1}}
\usepackage{subcaption}

\IEEEoverridecommandlockouts
 \newenvironment{cexmp}[2]
{\addtocounter{exmp}{-1}\begin{exmp}{\textit{continued} (#2)}}
  {\end{exmp}}

\newenvironment{sizeddisplay}[1]
 {\par\nopagebreak#1\noindent\ignorespaces}
 {\nopagebreak\ignorespacesafterend}
\graphicspath{ {figs/} }

\title{\LARGE \bf Decomposition-based Hierarchical  Task Allocation and Planning for Multi-Robots under Hierarchical Temporal Logic Specifications
}%
\author{Xusheng Luo$^{1}$, Shaojun Xu$^{2,\dagger}$, Ruixuan Liu$^{1}$ and Changliu Liu$^{1}$
\thanks{*This work was in part supported by NSF under grant No. 2144489.}
\thanks{$^{1}$Xusheng Luo, Ruixuan Liu and Changliu Liu are with Robotics Institute, Carnegie Mellon University, Pittsburgh, PA 15213, USA (e-mail: {\tt\small\{xushengl,  ruixuanl, cliu6\}@andrew.cmu.edu})}%
\thanks{{\color{black}$^{2}$Shaojun Xu with College of Control Science and Engineering, Zhejiang University, Hangzhou, 310058, China (e-mail: {\tt\small sjxu@zju.edu.cn}).}}%
\thanks{$^{\dagger}$Work done during internship at Carnegie Mellon University.}
}

\setboolean{arXiv}{true}  

\begin{document}

\maketitle

\begin{abstract}
Past research into robotic planning with temporal logic specifications, notably Linear Temporal Logic (LTL), was largely based on a single formula for individual or groups of robots. But with increasing task complexity, LTL formulas unavoidably grow lengthy, complicating interpretation and specification generation, and straining the computational capacities of the planners. A recent development has been the hierarchical representation of LTL~\cite{luo2024simultaneous} that contains multiple temporal logic specifications, providing a more interpretable framework. However, the proposed planning algorithm assumes the independence of robots within each specification, limiting their application to multi-robot coordination with complex temporal constraints. In this work, we formulated a decomposition-based hierarchical framework. At the high level, each specification is first decomposed into a set of atomic sub-tasks. We further infer the temporal relations among the sub-tasks of different specifications to construct a task network. Subsequently, a Mixed Integer Linear Program is used to assign sub-tasks to various robots. At the lower level, domain-specific controllers are employed to execute sub-tasks.  Our approach was experimentally applied to domains of navigation and manipulation. The simulation demonstrated that our approach can find better solutions using less runtimes.
\end{abstract}

\section{Introduction}
The field of temporal logic planning has seen rapid expansion in recent years, which can manage a more diverse range of tasks beyond the typical point-to-point navigation, encompassing temporal objectives as well. These tasks involve sequencing or coverage~\cite{fainekos2005temporal}, intermittent communication~\cite{kantaros2018distributed} and manipulation~\cite{he2015towards} among others. 
Here, we focus on Linear Temporal Logic (LTL) specifications~\cite{pnueli1977temporal}.  In the most research that centered on multi-robot systems, LTL tasks are either assigned {\it locally} to individual robots within a team~\cite{guo2015multi,tumova2016multi,yu2021distributed,bai2022hierarchical}, or a {\it global} LTL specification is used to capture the collective behavior of all robots. In the latter approach, tasks can either be explicitly assigned to each robot, as in~\cite{smith2011optimal,kantaros2020stylus,luo2021abstraction}, or task allocation is considered without specific assignments, as in~\cite{shoukry2017linear,schillinger2018simultaneous,sahin2019multirobot, leahy2021scalable,luo2022temporal,li2023fast}. However, except for~\cite{luo2024simultaneous}, {\color{black}all LTL formulas, regardless of whether they are assigned locally or globally, are in the ``flat'' structure, meaning a single LTL formula specifies the behavior of an individual robot or a team of robots. They tend to become cumbersome for complex tasks. Despite attempts to increase expressivity, such as inner and outer logics~\cite{sahin2019multirobot,liu2023robust}, they are task specific and still  flat formulas by syntactically putting all in single formulas.}

{\color{black}Studies (e.g.,~\cite{tenenbaum2011grow})} suggest that humans prefer hierarchical task specification, which improves interpretability of planning and execution, making it easier to identify ongoing work and conveniently adjust {\color{black}infeasible} parts without affecting other components. In our prior work \cite{luo2024simultaneous}, we introduced a hierarchical and interpretable structure for LTL specifications, consisting of various levels of flat specifications that enhance computational efficiency. A bottom-up planning approach was developed for simultaneous task allocation and planning. While this method is efficient, it operates under the assumption that  tasks with temporal dependencies within a flat specification is allocated to a single robot, hence robots can work independently, limiting its application in scenarios that require cooperation among multiple robots.

In this work, we developed a decomposition-based hierarchical framework capable of addressing multi-robot coordination under hierarchical temporal logic specifications. Temporal logic specifications that do not explicitly assign tasks to robots generally require decomposition to derive the necessary task allocation, which can be accomplished in three approaches: The most common method, used in works such as
\ifthenelse{\boolean{arXiv}}%
{%
\cite{schillinger2018simultaneous,luo2019transfer,camacho2019ltl,luo2022temporal,liu2024time},}{
\cite{schillinger2018simultaneous,luo2019transfer,camacho2019ltl,luo2022temporal,liu2024time},
}
involves decomposing a global specification into multiple tasks, leveraging the transition relations within the automaton, which represents an LTL formula graphically. Our work is also in line with this approach as graph-based methods are well-established. As demonstrated in~\cite{shoukry2017linear,sahin2019multirobot}, the second approach creates a Boolean Satisfaction or Integer Linear Programming (ILP) model, which simultaneously addresses task allocation and implicit task decomposition in a unified formulation. Another method, e.g.,~\cite{leahy2023rewrite}, directly interacts with the syntax tree of LTL formulas, which segments the global specification into smaller, more manageable sub-specifications. 

The proposed framework, built upon work~\cite{luo2022temporal}, starts at the top level by breaking down each flat specification into smaller sub-tasks. These sub-tasks are then organized into a {\it task network}, where we deduce the temporal relationships between them. Subsequently, we use a Mixed Integer Linear Programming (MILP) approach to assign these sub-tasks to various robots. At the lower level, we implement domain-specific planning methods to carry out these sub-tasks. {\color{black}To the best of knowledge, our work is the first one that can tackle multi-robot collaboration under hierarchical LTL specifications.} Our contributions are as follows:
\begin{enumerate}
\item We developed an efficient hierarchical planning algorithm that can handle multi-robot coordination under hierarchical LTL specifications;
\item {\color{black}The efficiency and the quality of solutions were demonstrated through navigation and manipulation tasks.}
\end{enumerate}

{\color{black} It is worth noting that hierarchical framework lacks completeness in two respects. Firstly, for the sake of completeness,~\cite{luo2022temporal} makes assumptions about the single automaton's structure, assumptions that might not hold across a set of automatons when examining relationships between hierarchical specifications. More importantly, without an adequate interface, the high-level task plans might prove infeasible by the low-level controllers.}

\section{Problem Formulation}\label{sec:preliminaries}
\subsection{Linear Temporal Logic}
 {\color{black}Linear Temporal Logic (LTL) is composed of a set of atomic propositions $\mathcal{AP}$,} along with boolean operators such as conjunction ($\wedge$) and negation ($\neg$), as well as temporal operators like next ($\bigcirc$) and until ($\mathcal{U}$)~\cite{baier2008principles}. LTL formulas follow the syntax outlined below:
\begin{align}\label{eq:grammar}
\phi:=\top~|~\pi~|~\phi_1\wedge\phi_2~|~\neg\phi~|~\bigcirc\phi~|~\phi_1~\mathcal{U}~\phi_2, 
\end{align}
where $\top$ is an unconditionally true statement, and $\pi\in \mathcal{AP}$ refers to a boolean valued atomic proposition. Other temporal operators can be derived from $\mathcal{U}$, such as $\Diamond \phi$ that implies $\phi$ will ultimately be true at a future time.

\ifthenelse{\boolean{arXiv}}%
{%
An infinite {\it word} $w$ over the alphabet $2^{\mathcal{AP}}$, where $\mathcal{AP}$ is the set of atomic propositions, can be denoted as $w=\sigma_0\sigma_1\ldots\in (2^{\mathcal{AP}})^{\omega}$, with $\omega$ signifying infinite repetition, and $\sigma_k\in2^{\mathcal{AP}}$ for $\forall k\in\mathbb{N}$. The {\it language} $\texttt{Words}(\phi)$ is the collection of words that meet the formula $\phi$. This means, $w$ belongs to $\texttt{Words}(\phi)$ if and only if $w$ satisfies $\phi$.}
{
}

We focus on a subset of LTL known as syntactically co-safe formulas, or sc-LTL for short~\cite{kupferman2001model}. It has been established that any LTL formula encompassing only the temporal operators $\Diamond$ and $\mathcal{U}$ and written in positive normal form (where negation is exclusively before atomic propositions) is classified under syntactically co-safe formulas~\cite{kupferman2001model}. Sc-LTL formulas can be satisfied by finite sequences followed by any infinite repetitions. {\color{black}An LTL formula $\phi$ can be translated into a Nondeterministic B$\ddot{\text{u}}$chi Automaton (NBA)~\cite{baier2008principles}}:
\begin{defn}[NBA]\label{def:nba}
 { An NBA $B$ is  a tuple $B=\left(\ccalQ, \ccalQ_0,\Sigma,\rightarrow_B,\mathcal{Q}_F\right)$, where $\ccalQ$ is the set of states; $\ccalQ_0\subseteq\ccalQ$ is a set of initial states; $\Sigma=2^{\mathcal{AP}}$ is an alphabet;  $\rightarrow_B\subseteq\ccalQ\times \Sigma\times\ccalQ$ is the transition relation;
and $\ccalQ_F\subseteq\ccalQ$ is a set of accepting states.}
\end{defn}

\subsection{Hierarchical Linear Temporal Logic}
{\color{black}\begin{defn}[Hierarchical LTL~\cite{luo2024simultaneous}]\label{def:hltl} A hierarchical LTL specification, denoted by $\Phi = \{\level{i}{k}  \mid k = 1, \ldots,K, i = 1, \ldots, |\Phi^k|\}$ where $\level{i}{k}$ is the $i$-th sc-LTL specification at level $k$, $\Phi^k$ denotes the set of specifications at level $k$, and $|\cdot|$ denotes the cardinality, includes $K$ levels such that each specification at level $k$, for $k = 1, \ldots, K-1$, is constructed from specifications at the immediate lower level $k+1$.
\end{defn}

We refer to each specification $\level{i}{k}$ in $\Phi$ as the ``flat'' specification.}
These flat specifications can be organized in a tree-like specification hierarchy graph, where each node represent a flat sc-LTL specification. Edges between nodes indicate that one specification encompasses another as a {\it composite proposition}. This composite proposition is, in essence, another flat sc-LTL formula. Leaf nodes represent {\it leaf specifications} at the $K$-th level that consist only of atomic propositions, while non-leaf nodes represent {\it non-leaf specifications} made up of composite propositions. 
\begin{exmp}[Multi-Robot Pickup and Delivery (MRPD)]\label{exmp:mrpd}
Consider a MRPD problem in a warehouse setting. Here, mobile robots, waiting at the dock, are assigned to pick up various items from storage shelves and deliver them to a packing zone. The warehouse is segmented into six areas: grocery, health, outdoor, pet supplies, furniture, and electronics; see Fig.~\ref{fig:supermarket}. We consider three types of robots. There are two robots for each of these types. The specific requirements for task 1 are as follows: 1) {\it Initially}, a robot of type 1 proceeds to the furniture section and waits {\it until} the arrival of a type 3 robot to assist with the loading of a large piece of furniture; {\it following} this, it gathers items from the outdoor and pet sections {\it in no particular sequence}. 2) A type 2 robot should {\it initiate} its task by gathering items from the health section, and {\it only afterward} does it move to the grocery section, ensuring it doesn't visit the grocery section {\it before} visiting the health section. 3) A type 3 robot moves to the furniture section to assist the type 1 robot. 4) {\it After} the items are delivered to the packing area, all robots {\it eventually} return to the dock.  Temporal constraints may arise from item attributes like weight, fragility, and category. 
\ifthenelse{\boolean{arXiv}}%
{%
This example extends the classical MRPD in several aspects: (i) Tasks are defined within temporal logic specifications, not as predefined point-to-point navigation. (ii) Logical and temporal constraints between tasks, not considered in typical MRPD, are included. (iii) A single robot may handle multiple tasks, contrasting with most MRPD works where each robot performs one task, except for~\cite{chen2021integrated,xu2022multi}. (iv) Cooperative behaviors among robots are introduced but not typically included in MRPD.
}{%
}
\begin{figure}[!t]
    \centering
    \includegraphics[width=\linewidth]{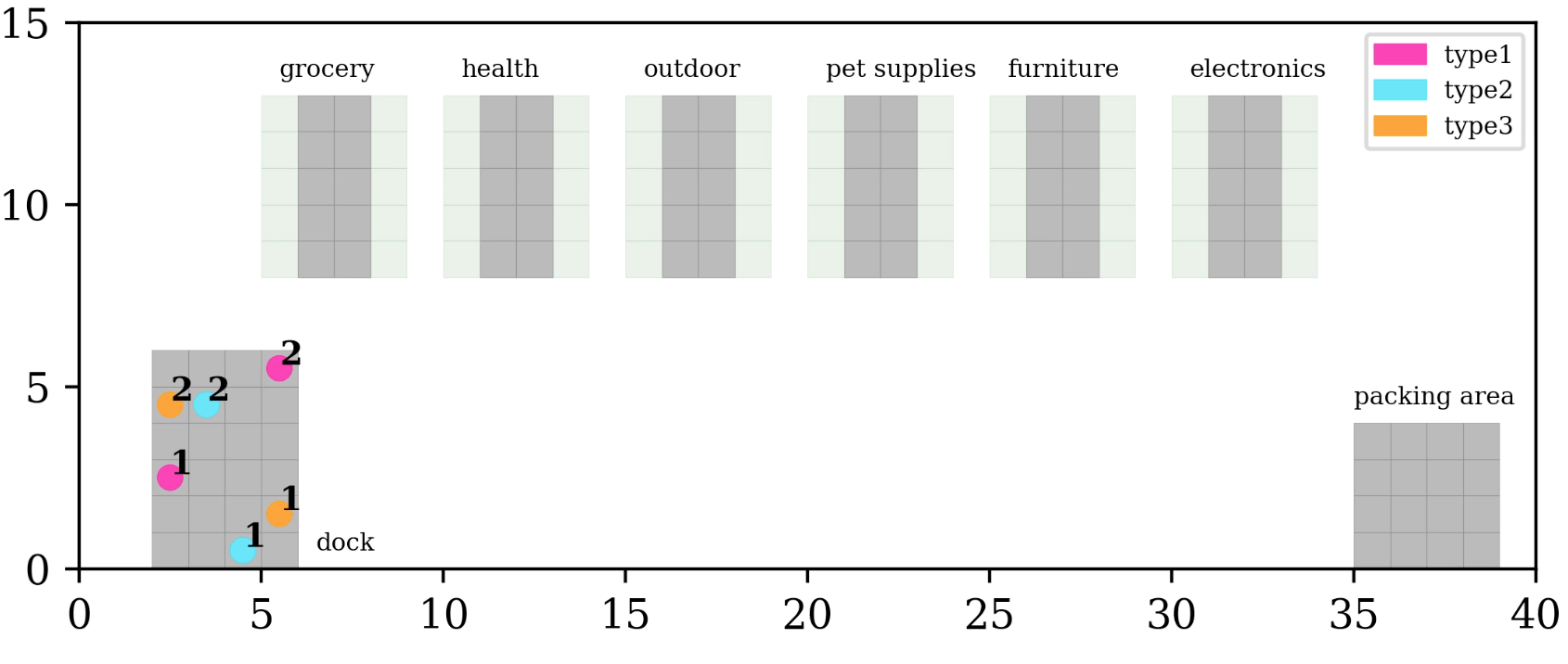}
    \caption{Topological map of the supermarket. Various types of robots are represented in distinct colors.}
    \label{fig:supermarket}
\end{figure}

{\color{black}Let symbols $\pi_{s}^{t}$ represent atomic propositions indicating the need for a robot of type $t$ to collect items from section $s$. Propositions with the same second superscript, like $\pi_{s}^{t,j}$ and $\pi_{s'}^{t,j}$, must be executed by the same robot of type $t$. For instance, $\apsm{furn}{1}{1}, \apsm{outd}{1}{1}$ and $\apsm{pet}{1}{1}$ should be completed by the same robot of type 1.} The hierarchical specification $\Phi$ is:
\begin{sizeddisplay}{\small}
\begin{align}\label{eq:mrpd}
L_1: \quad & \level{1}{1} = \Diamond \level{1}{2} \wedge \Diamond \level{2}{2} \nonumber \\
L_2: \quad & \level{1}{2} = \Diamond \left(\level{1}{3} \wedge \Diamond  \level{2}{3} \wedge \Diamond (\level{3}{3} \wedge \Diamond \level{4}{3})\right) \nonumber \\
     &  \level{2}{2}  = \Diamond (\level{5}{3} \wedge \Diamond \level{6}{3}) \nonumber \\
L_3: \quad & \level{1}{3} = \Diamond (\pi_{\text{furn}}^{1,1} \wedge \bigcirc (\pi_{\text{furn}}^{1,1} \; \mathcal{U}\; \pi_{\text{furn}}^{3,3}))  \nonumber \\
& \level{2}{3} =  \Diamond (\pi_{\text{pack}}^{3,3} \wedge \Diamond \pi_{\text{dock}}^{3,3}) \\
&  \level{3}{3}= \Diamond \pi_{\text{outd}}^{1,1} \wedge \Diamond \pi_{\text{pet}}^{1,1} \nonumber\\
&  \level{4}{3} = \Diamond (\pi_{\text{pack}}^{1,1} \wedge \Diamond \pi_{\text{dock}}^{1,1}) \nonumber\\
&  \level{5}{3} = \Diamond \pi_{\text{heal}}^{2,2} \wedge \Diamond \pi_{\text{groc}}^{2,2} \wedge  \neg  \pi_{\text{groc}}^{2,2}  \;  \mathcal{U}\; \pi_{\text{heal}}^{2,2} \nonumber \\
&  \level{6}{3} = \Diamond (\pi_{\text{pack}}^{2,2} \wedge \Diamond \pi_{\text{dock}}^{2,2}). \nonumber
\end{align}
\end{sizeddisplay}
The specification hierarchy graph is shown in Fig.~\ref{fig:shg}. 
\ifthenelse{\boolean{arXiv}}%
{%
When $\level{i}{k}$ appears  at the right side of a formula,  it is referred to as a composite proposition, while at the left side, it's a specification.
For instance, in the formula $\level{2}{2} = \Diamond (\level{5}{3} \wedge \Diamond \level{6}{3})$, $\level{2}{2}$ is a specification, and $\level{5}{3}$ and $\level{6}{3}$ are composite propositions.
A flat LTL specification of this task can be seen in Eq.~\eqref{eq:flat_task1} in Appendix~\ref{app:task}.  Note that the flat and hierarchical versions of this task are not strictly equivalent in terms of the set of accepted words.  In the flat form, depicting precedence often necessitates using the sub-formula $\neg \pi_a \, \mathcal{U}\, \pi_b$ frequently, especially due to the multitude of atomic propositions and the mix of independent and dependent relations. However, in hierarchical LTL, specifications are typically short, allowing for a simpler expression of precedence using $\Diamond$, which is less rigid than $\mathcal{U}$. Nevertheless, either form adheres to task descriptions in natural language, as these descriptions usually don't express a preference for a specific type of precedence relation.\hfill $\square$
}{%
A flat LTL specification of this task can be seen in~\cite{luo2023robotic}. \hfill $\square$
}

\begin{figure}[!t]
    \centering
    \includegraphics[width=1\linewidth]{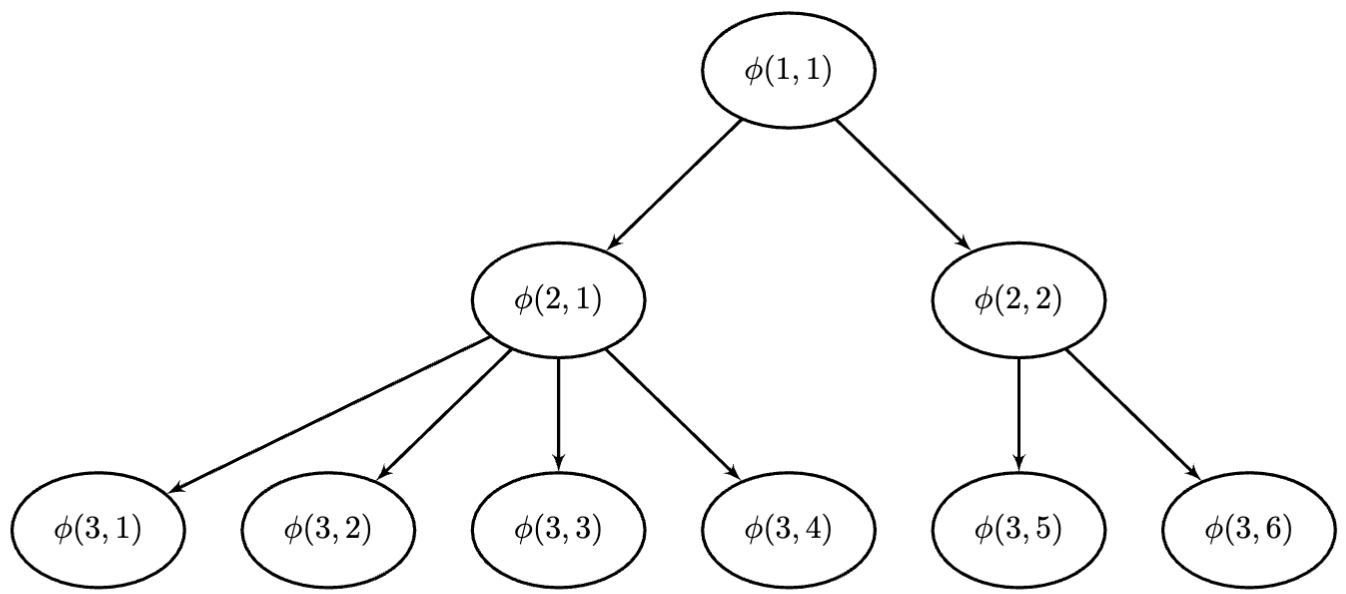}
\caption{Specification hierarchy graph for Example~\ref{exmp:mrpd}.}
    \label{fig:shg}
\end{figure} 

\end{exmp}

\subsection{Problem Formulation}\label{sec:problem}
{\color{black}\begin{problem}\label{prob:hltl}
Consider an environment $\mathcal{E}$ populated by a set of heterogeneous robots $\mathcal{R} = \{r_1, ..., r_n\}$ of different capabilities, and a task represented by a hierarchical LTL specification $\Phi$ for either navigation or manipulation. The goal is to generate a plan $\mathcal{P}$ that allocates tasks to robots and schedules actions in a way that fulfills specification $\Phi$, provided that such a plan is feasible.
\end{problem}

\begin{rem}
This work primarily addresses the high-level task allocation under hierarchical specifications. We assume that robot dynamics and their interactions with the environment are handled by domain-specific low-level controllers.
\end{rem}}
\section{Decomposition-based Hierarchical Planning}
Our approach builds upon work~\cite{luo2022temporal}, which decomposes the NBA of a given LTL specification into sub-tasks (which will be defined in Definition~\ref{defn:sub-task}) and infers their temporal relationships. Next, we  deduce the temporal relations between atomic sub-tasks across all leaf specifications.

We start by presenting the necessary definitions and notations related to NBA. {\color{black}It is} worth noting that the NBA, as defined in Definition~\ref{def:nba}, can be perceived as a graph. For the sake of simplicity,  {\color{black}we will} refer to the NBA as the graph \( \autop \). When discussing \( \autop \)'s edges, we don't consider self-loops as they can be represented by vertices themselves. The propositional formula \( \gamma \) associated with a transition \( v_1 \xrightarrow{\gamma} v_2 \) in the NBA \( \autop \) is called a {\it vertex label} if \( v_1 = v_2 \), and an {\it edge label} if not. We denote the mappings of a vertex and an edge to their respective labels using the functions $\gamma: \ccalV \to \Sigma$ and $\gamma: \ccalV \times \ccalV \to \Sigma$, respectively. We then define an edge-induced {\it sub-task} as a series of actions that robots must perform to trigger a transition in the NBA. 
\ifthenelse{\boolean{arXiv}}%
{To greatly decrease the number of edge-induced sub-tasks in the NBA, we remove any edge where the label requires that more than one proposition be true simultaneously, if such a condition is not required by the corresponding specification.}{
}

\begin{defn}[\color{black}{Edge-induced sub-task~\cite{luo2022temporal}}]\label{defn:sub-task}
  Given an edge $(v_1, v_2)$ in $\autop$, an edge-induced sub-task is defined by the edge label $\gamma(v_1, v_2)$ and the starting vertex label $\gamma(v_1)$.
\end{defn}

\begin{defn}[Atomic and composite sub-task]\label{defn:atomic_comp_sub-task}
In the NBA $\ccalA_\phi$, a sub-task $(v_1, v_2)$ is classified as an atomic sub-task if the edge label $\gamma(v_1, v_2)$ consists solely of atomic propositions. Otherwise, {\color{black}it is} a composite sub-task if the edge label consists solely of composite propositions.
\end{defn}

\begin{figure}
\begin{minipage}[c][6.2cm][t]{.55\linewidth}
  \vspace*{\fill}
  \centering
    \subfigure[NBA of $\level{1}{1}$]{
      \label{fig:par_a} 
       \includegraphics[width=1.03\linewidth]{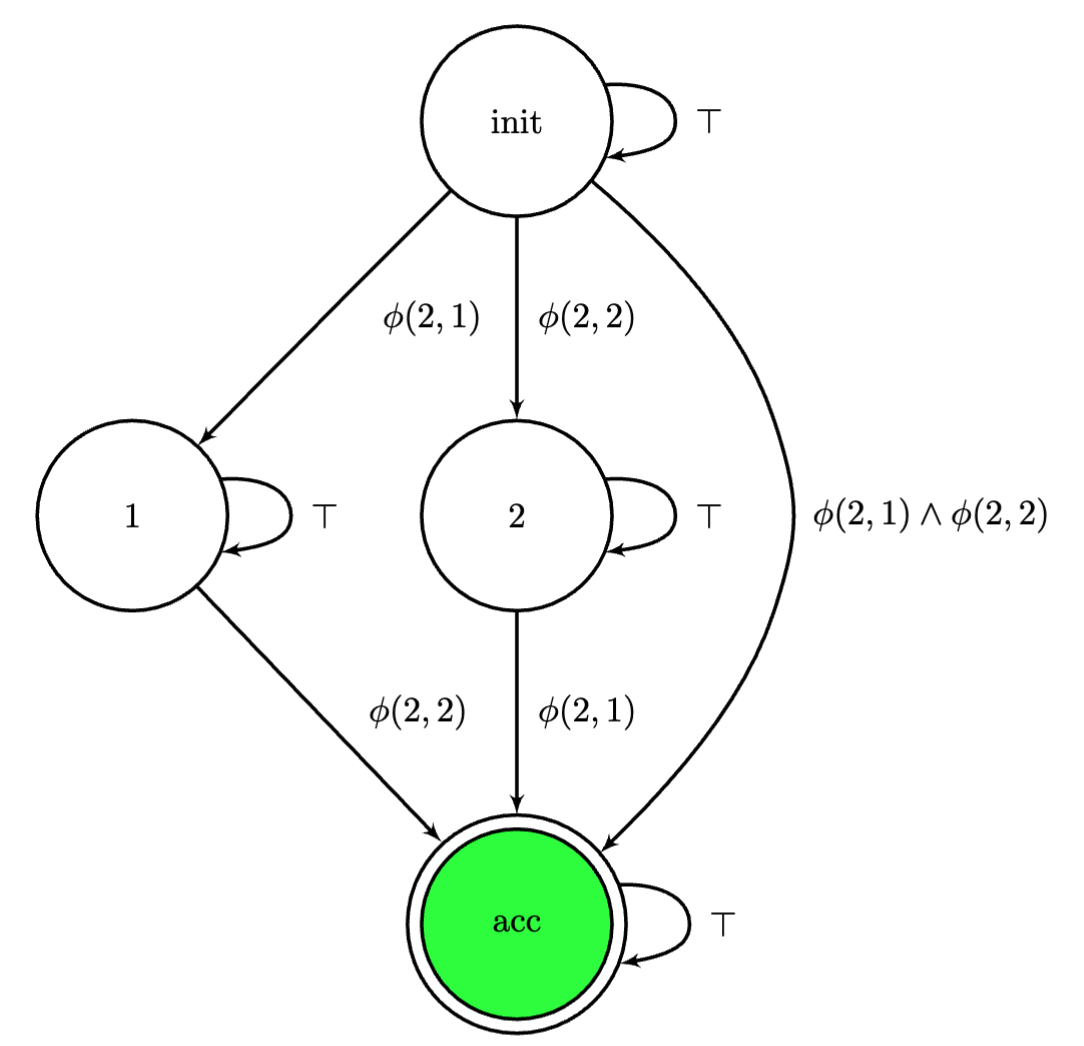}} \\
\end{minipage}%
\begin{minipage}[c][5cm][t]{.35\linewidth}
  \subfigure[Sub-tasks]{
      \label{fig:par_b} 
      \includegraphics[width=0.9\linewidth]{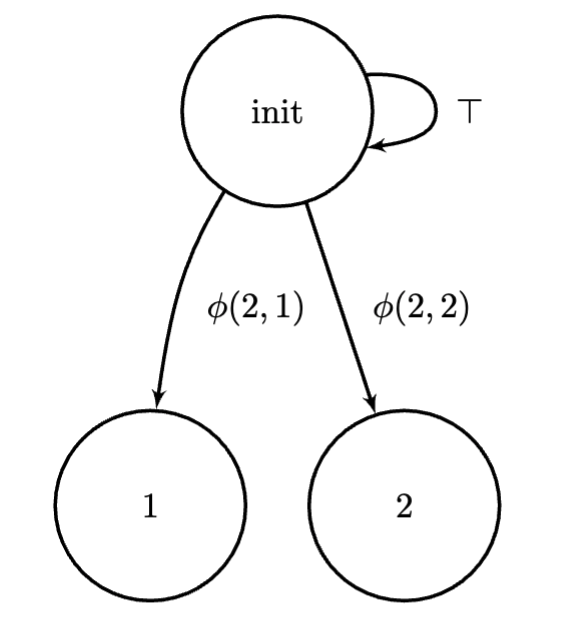}} \\
     \subfigure[Independent sub-tasks]{
      \label{fig:par_c}
      \includegraphics[width=1.2\linewidth]{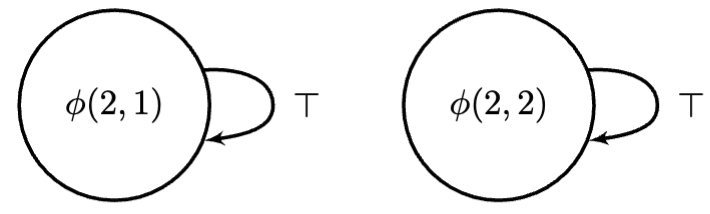}}
\end{minipage}
\caption{(a) The NBA of specification $\level{1}{1}$ in Eq. \eqref{eq:mrpd}. The sub-task $(v_{\text{init}}, v_1)$ implies that the composite proposition $\level{1}{2}$ must eventually be true, as indicated by its edge label. The sub-tasks $(v_{\text{init}}, v_1)$ and $(v_2, v_{\text{acc}})$ are equivalent because $\gamma(v_{\text{init}})=\gamma(v_2)$ and $\gamma(v_{\text{init}}, v_1)= \gamma(v_2, v_{\text{acc}})$. Similarly, the sub-tasks $(v_{\text{init}}, v_2)$ and $(v_1, v_{\text{acc}})$ are equivalent. (b) The two sub-tasks $(v_{\text{init}}, v_1)$ and $(v_{\text{init}}, v_2)$. (c) Concise representation of sub-tasks. The two sub-tasks are temporally independent, thus no edges exist between them. }
\label{fig:par}
\end{figure}

Two sub-tasks are {\it equivalent} if they have identical starting vertex labels and edge labels. Note that because each specification consists solely of atomic propositions (for leaf specifications) or composite propositions (for non-leaf specifications), all of its sub-tasks should be of the same type.
 
\begin{cexmp}{eg:lego}{Sub-tasks}\label{exmp:mrta-!}
The NBA of specification $\level{1}{1}$ and sub-tasks are shown in Figs.~\ref{fig:par_a} and~\ref{fig:par_b}.
\end{cexmp}

\subsection{Generate Temporal Relations across Atomic Sub-tasks}
{\color{black}The work~\cite{luo2022temporal} focuses on a flat LTL specification, extracting edge-induced sub-tasks from its NBA, and identifying the temporal relationships among these sub-tasks, i.e., a sub-task must be completed before, after or independently of another one;} an example with specification $\level{1}{1}$ of task 1 is illustrated in Fig.~\ref{fig:par_c}. The goal here is to create a task network represented as a Directed Acyclic Graph (DAG). In this DAG, each node represents an atomic sub-task from a specific leaf specification, and edges indicate precedence relationships between atomic sub-tasks, as determined by corresponding non-leaf specifications.

Let $\Phi^k$ denote the set of specifications at the $k$-th level, with $|\Phi^k|$ indicating its size. In what follows, we denote the edge-induced sub-task as $e = (v_1, v_2)$ and use $\tasksall$, $\tasks{\phi}{a}$, and $\tasks{\phi}{c}$ to represent the sets of all sub-tasks, atomic sub-tasks, and composite sub-tasks for a given specification $\phi \in \Phi$, respectively. As defined in Def.~\ref{defn:atomic_comp_sub-task}, $\tasksall$ equates either to $\tasks{\phi}{a}$ or $\tasks{\phi}{c}$. For any two different sub-tasks $e, e' \in \tasksall$, we use $e \bowtie e'$ to denote that sub-task $e$ must be completed before ($\prec$), after ($\succ$), or independently ($\|$) of $e'$. The following assumption {\color{black}restricts} the relationships between sub-tasks from two composite propositions.
\begin{asmp}[Temporal relation inheritance]\label{asmp:inheritance}
Sub-tasks follow the temporal relations of corresponding specifications. That is, if $\phi \bowtie \phi'$, then $ e \bowtie e', \forall e \in \tasks{\phi}{}, \forall e' \in \tasksallp$.
\end{asmp}

\begin{algorithm}[!t]
   \caption{Construct DAG $\ccalG = (\ccalV, \ccalE)$}
   \LinesNumbered
   \label{alg:global_tem_rela}
   \KwIn {Hierarchical LTL specifications $\{\level{i}{k}\}_{k=1}^K$}
   \KwOut {DAG $\ccalG$ of atomic sub-tasks}
   $\ccalV = \varnothing, \ccalE = \varnothing$\;
   \Comment*[r]{{\color{black}Same leaf specifications}}
    \For{$i \in \{1, \ldots, |\Phi^K|\}$\label{dag:spec}}{
        \For{$e, e' \in \tasks{\level{i}{K}}{a}$\label{dag:same_spec_start}}{
        \textbf{if} {$e \prec e'$} \textbf{then} \text{$\ccalE = \ccalE \cup \{(e, e')\}$}\;
        \textbf{if} {$e \succ e'$} \textbf{then} \text{$\ccalE = \ccalE \cup \{(e', e)\}$ \label{dag:same_spec_end}}\;
        $\ccalV = \ccalV \cup \{e, e'\}$\;
        }
    }
    \Comment*[r]{{\color{black}Different leaf specifications}}
    \For{$i, i' \in \{1, \ldots, |\Phi^K|\}$\label{dag:inter_spec_start}}{
            \textbf{if} {$i == i'$} \textbf{then} \textbf{continue}\;
            Get the minimal common predecessor $\mu_{cp}(i, i')$\; \label{dag:pmin}
            Find sub-tasks $e_\tphi$ and $e_{\tphi'}$ in $\mu_{cp}(i, i')$ whose edge labels include $\phi(K, i)$ and $\phi(K, i')$, resp.\; \label{dag:diff_parent_start}
            \For{$e \in \tasks{\level{i}{K}}{a}$\label{dag:for_start}}{
                \For{$e' \in \tasks{\level{i'}{K}}{a}$}{
                \textbf{if} {$e_{\tphi} \prec e_{\tphi'}$} \textbf{then} \text{$\ccalE = \ccalE \cup \{(e, e')\}$}\;
                \textbf{if} {$e_{\tphi} \succ e_{\tphi'}$} \textbf{then} \text{$\ccalE = \ccalE \cup \{(e', e)\}$\label{dag:inter_spec_end} \label{dag:diff_parent_end}}\;
                }
            }
    }
    \textbf{return} $\ccalG = (\ccalV, \ccalE)$\;
 \end{algorithm}

The process for creating the DAG $\ccalG = (\ccalV, \ccalE)$ is detailed in Alg.~\ref{alg:global_tem_rela}, which consists of two parts, based on whether the atomic sub-tasks belong to the same leaf specification.

\subsubsection{Same leaf specification ${\level{i}{K}}$} The algorithm iterates through each leaf specification $\level{i}{K}$. For any pair of atomic sub-tasks within $\tasks{\level{i}{K}}{a}$, an edge is added to the graph $\ccalG$ if they have a precedence relation [lines \ref{dag:spec}-\ref{dag:same_spec_end}]. The method for extracting temporal relations between sub-tasks is based on the approach described in Section IV.C of~\cite{luo2022temporal}.

\subsubsection{Different leaf specifications ${\level{i}{K}}$ and ${\level{i'}{K}}$} For a given specification $\level{i}{K}$, its {\it predecessor} is defined as the composite proposition that encompasses it at the higher level. The sequence of predecessors leading up to the root specification, denoted as $\ccalS(\level{i}{K})$, is established by tracing back to the first level in the specification hierarchy graph. Note that the sequence of predecessors also includes the specification itself. When examining two sequences $\seqans{i}{K}$ and $\seqans{i'}{K}$, let the set $\Phi_\wedge(i, i') = \left\{  \phi \, | \, \phi \in \ccalS(\level{i}{K}), \phi \in \ccalS(\level{i'}{K}) \right\}$ collect predecessors that appear in both sequences. 
{\color{black}A predecessor is defined as the {\it minimal common predecessor}, denoted by $\mu_{cp}(i, i')$, if it is the closest to leaf specifications, that is, $
  \mu_{cp}(i, i') = \argmin_{\phi \in \Phi_\wedge (i, i') } \texttt{dist}( \phi - \level{i}{K})$,
where $\texttt{dist}$ returns the distance between two specifications in terms of the number of levels [line~\ref{dag:pmin}].} The temporal relations between the sets $\tasks{\level{i}{K}}{a}$ and $\tasks{\level{i'}{K}}{a}$ are determined within this minimal common predecessor. It may happen that $\level{i}{K}$ or $\level{i'}{K}$ is not directly included in $\mu_{cp}(i, i')$ due to multiple levels separating them. In such cases, we identify the composite propositions within $\mu_{cp}(i, i')$, labelled as $\tphi$ and $\tphi'$, which act as predecessors to $\level{i}{K}$ and $\level{i'}{K}$, respectively [line~\ref{dag:diff_parent_start}].  Subsequently, we identify a composite sub-task $e_\tphi$ within $\tasks{\mu_{cp}(i, i')}{c}$ that contains $\tphi$ in its edge label, without being preceded by the negation operator $\neg$. This sub-task necessitates fulfilling the composite proposition $\tphi$, which in turn requires satisfying $\level{i}{K}$. Similarly, we find another composite sub-task $e_{\tphi'}$ corresponding to $\level{i'}{K}$ in $\tasks{\mu_{cp}(i, i')}{c}$. The temporal relationship between $\level{i}{K}$ and $\level{i'}{K}$ mirrors that between their associated sub-tasks $e_\tphi$ and $e_{\tphi'}$ within $\mu_{cp}(i, i')$. As per Assumption \ref{asmp:inheritance}, this relationship extends to the atomic sub-tasks in $\tasks{\level{i}{k}}{a}$ and $\tasks{\level{i'}{k'}}{a}$; see lines \ref{dag:for_start}-\ref{dag:diff_parent_end}.

\begin{figure}[!t]
    \centering
    \includegraphics[width=1\linewidth]{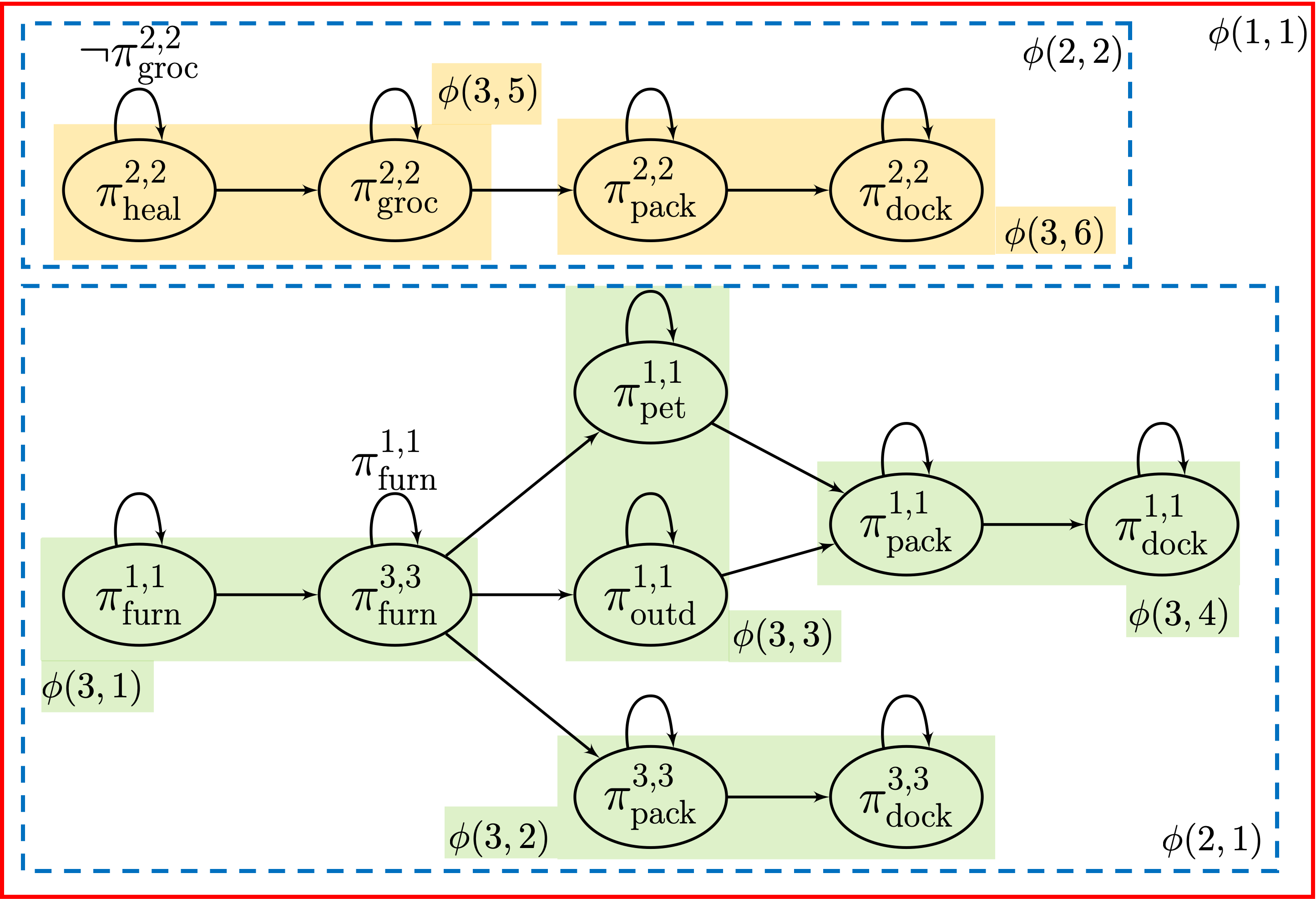}
    \caption{Task network of task 1. {\color{black}Each sub-task is represented by a node along with its self-loop. The symbol within each node corresponds to the edge label of the sub-task. The label $\top$ associated with self-loops is omitted.} Nodes with sub-tasks belonging to the same leaf specification are grouped within a shaded block, with the leaf specification being marked nearby. {\color{black}Sub-tasks associated with $\level{1}{2}$ and $\level{2}{2}$ are encased in blue boxes, while those pertaining to the root specification $\level{1}{1}$ are enclosed in the red box.}}
    \label{fig:task_network}
\end{figure} 
 
\begin{cexmp}{eg:lego}{Temporal relations}\label{exmp:tem} {\color{black}The task network of task 1 is shown in Fig.~\ref{fig:task_network}.} In terms of the specifications $\level{1}{3}$ and $\level{5}{3}$, sequences of predecessors are $\ccalS(\level{1}{3}) = \{\level{1}{3}, \level{1}{2}, \level{1}{1}\},  \ccalS(\level{5}{3}) = \{\level{5}{3}, \level{2}{2}, \level{1}{1} \}$, minimal common predecessor is $\mu_{cp}(1,5) = \level{1}{1}$. 
Note that $\level{1}{3}$ and $\level{5}{3}$ do not fall within $\mu_{cp}(1,5)$ because they are separated by two levels. Predecessors in $\mu_{cp}(1, 5)$ are $\tphi=\level{1}{2}$ and $\tphi'=\level{2}{2}$, respectively; see Fig.~\ref{fig:shg}. Related composite sub-tasks in $\tasks{\mu_{cp}(1, 5)}{c}$ are $e_\tphi = (v_{\text{init}}, v_1)$ and $e_{\tphi'} = (v_{\text{init}}, v_2)$, respectively. Since they are temporally independent, we have $\level{1}{2}\, \|\, \level{2}{2}$.  By denoting atomic sub-tasks using their respective edge labels, the temporal relationships between atomic sub-tasks from the two specifications $\level{1}{3}$ and $\level{5}{3}$ are $\apsm{furn}{1}{1} \, \|\,  \apsm{heal}{2}{2}, \apsm{furn}{1}{1} \, \|\,  \apsm{groc}{2}{2}, \apsm{furn}{3}{3} \, \|\,  \apsm{heal}{2}{2}, \apsm{furn}{3}{3} \, \|\,  \apsm{groc}{2}{2}$; see the blocks associated with $\level{1}{3}$ and $\level{5}{3}$ in Fig.~\ref{fig:task_network}.
\end{cexmp}

\subsection{Formulate MILP to Allocate Sub-tasks}
Using the task network $\ccalG$, we adapt the MILP formulation from~\cite{luo2022temporal} to assign atomic sub-tasks to robots, which was inspired by the Vehicle Routing Problem (VRP) with temporal constraints~\cite{bredstrom2008combined}. The difference is that, for the case of a single specification, all sub-tasks must be completed. However, in this work, sub-tasks originate from various leaf specifications and not all are mandatory. For example, in a case where two leaf specifications are linked by an \texttt{OR} operator in a non-leaf specification (e.g., $\level{1}{1} = \level{1}{2} \vee \level{2}{2}$), {\color{black}it is} possible for one leaf specification, like $\level{1}{2}$, to be false, so all its sub-tasks should not be fulfilled. The MILP modifications are tailored to accommodate specification-level constraints, involving two categories of constraints: logical constraints and temporal constraints.  The solution provides a time-stamped sequential plan including the starting time (when the vertex label is activated) and the completion time (when the edge label becomes true) of  sub-tasks within $\ccalG$. 

For any specification $\phi \in \Phi$, considering a sub-task $(v_1, v_2) \in \tasksall$, let $\gamma$ be the propositional formula associated with its  edge or vertex label, i.e., $v_1 \xrightarrow{\gamma} v_2$ or $v_1 \xrightarrow{\gamma} v_1$. Assume this formula is in {\it disjunctive normal form} (DNF), i.e., $\gamma = \bigvee_{p\in \ccalP} \bigwedge_{q\in \ccalQ_p} (\neg) \pi$ or $\gamma = \bigvee_{p\in \ccalP} \bigwedge_{q\in \ccalQ_p} (\neg) \phi'$,  where the negation operator only precedes the atomic propositions $\pi$ or composite propositions $\phi'$, and $\ccalP$ and $\ccalQ_p$ are suitable index sets. Every propositional formula has an equivalent DNF form~\cite{baier2008principles}. {\color{black}We define $\ccalC_p^{\gamma}=\bigwedge_{q\in \ccalQ_p} (\neg) \pi$ or $\ccalC_p^{\gamma}=\bigwedge_{q\in \ccalQ_p} (\neg) \phi'$ as the $p$-th {\it clause} of $\gamma$ which includes a set $\ccalQ_p$ of {\it positive} and {\it negative} literals with each literal being an atomic or composite proposition.} 

\subsubsection{{Logical} Constraints}
 Logical constraints mainly encode the logical relation between labels, clauses and literals, and the realization of literals. We iterate over each specification $\phi \in\Phi$. Consider a specification $\phi \in\Phi$, for any edge or vertex label $\gamma$ within its NBA, the necessity to satisfy $\gamma$ depends on whether the specification $\phi$ should be true. To incorporate this condition, we introduce a binary variable $b_{\phi}$, representing the truth of specification $\phi$.  By default, the binary variable for the root specification is assigned a value of 1, i.e., $b_{\level{1}{1}} = 1$, which states that the root specification needs to be satisfied. The condition that $\gamma$ should be true if and only if the specification $\phi$ is true can be expressed by
\begin{equation}\label{eq:logic_task}
    \sum_{p \in \ccalP} b_p = b_{\phi}, \quad \forall  \phi\in \Phi,
\end{equation}
{\color{black}where the binary variable $b_p$ denotes  whether the $p$-th clause is used for the satisfaction of $\gamma$. If $b_{\phi} = 1$, constraint~\eqref{eq:logic_task} states that exactly one clause is used to satisfy $\gamma$.} Furthermore, the truth of a clause determines the truth of positive and negative literals inside it. In cases where $\phi$ is a non-leaf specification, all literals are essentially composite propositions. 
The condition that every positive literal must be true if its associated clause is true is formulated as
\begin{equation}\label{eq:comp}
    \sum_{\phi \in \mathsf{lit}^+(\ccalC_p^{\gamma})} b_{\phi} = \left|\mathsf{lit}^+(\ccalC_p^{\gamma})\right| b_p, \quad \forall p \in \ccalP,
\end{equation}
{\color{black}where $\mathsf{lit}^+(\ccalC_p^{\gamma})$ is the set of positive literals in the $p$-th clause}.  Beginning with the root specification, which must be satisfied ($b_{\level{1}{1}}=1$), the satisfaction of composite propositions within $\level{1}{1}$, i.e., specifications at level 2, is deduced according to Eq.~\eqref{eq:logic_task}. Their truth will influence the satisfaction of level 3 specifications, and this procedure continues down to the level of leaf specifications. At this point, only those atomic sub-tasks linked to leaf specifications identified as satisfied are assigned and executed. This assignment and execution process is governed by the following set of inequality constraints.

\subsubsection{{Temporal} Constraints}
These constraints capture the temporal orders between atomic sub-tasks, which are represented as linear inequalities. The key idea here is that if the specification $\phi$ should not be satisfied, the inequality constraints associated with it should be trivially met. Each of these constraints is represented by a general linear inequality $g(x) \leq 0$, where $x$ denotes the decision variables of the MILP. Let $\Phi_g$ denote the set of leaf specifications involved in $g(x)$. The following constraint ensures that $g(x)$ is trivially satisfied when at least one leaf specification in $\Phi_g$ is false:
\begin{align}
    g(x) \leq \sum_{\phi \in \Phi_g } M (1 - b_{\phi}),
\end{align}
where $M$ represents a sufficiently large integer. The inequality is only active when $b_\phi=1, \forall \phi \in \Phi_g$. For differentiation, the notation \( \langle \cdot \rangle \) is used when referencing constraints in~\cite{luo2022temporal}. Consider constraint $\langle 15 \rangle$ stating that the completion time of an atomic sub-task $e \in \tasks{\phi}{a}$ in the task network should be later than that of atomic sub-task $e'\in \tasks{\phi'}{a}$ if $e'$ is required to be completed before $e$. Let $t_e$ be an integer variable denoting the completion time of sub-task $e$, that is, when its edge label is satisfied. Now the constraint becomes 
\begin{align}\label{eq:completion_time}
    t_{e'} +1  \leq t_e + M (1 - b_{\phi}) & + M (1 - b_{\phi'}), \quad  \text{if} \; e' \prec e.
\end{align}
The constraint~\eqref{eq:completion_time} comes into effect, i.e., $ t_{e'} +1  \leq t_e $, only if both $\phi$ and $\phi'$ need to be true. 
\ifthenelse{\boolean{arXiv}}%
{%
The same logic applies to the other constraints $\langle 13 \rangle$, $\langle 17 \rangle$, $\langle 18 \rangle$, $\langle 19 \rangle$, $\langle 23 \rangle$ and $\langle 25 \rangle$, which are elaborated in Appendix~\ref{app:milp}.
}{%
The modifications to other constraints along with objectives are detailed in Appendix~\ref{app:milp}.
 }
 
\begin{rem}
For navigation tasks, a sequence of Generalized Multi-Robot Path Planning (GMRPP) problems~\cite{luo2022temporal}, are set up for each sub-task to create robot paths. These GMRPP solutions are sequentially organized and evaluated to ensure they align with the semantics of hierarchical LTL specifications~\cite{luo2024simultaneous}. If this satisfaction check fails, no solution is presented, resulting in the method's soundness. {\color{black}It is} worth noting that~\cite{luo2022temporal} offers a complete solution for a wide range of LTL specifications. Leveraging this foundation, our approach is feasible for many robotic tasks, a claim supported by the following experiments.
\end{rem}

 \section{Numerical Experiments}\label{sec:sim}
\ifthenelse{\boolean{arXiv}}%
{%
We use Python 3.10.12 on a computer with 3.5 GHz Apple M2 Pro and 16G RAM. The Big-M based MILP is {\color{black}solved} using the Gurobi solver~\cite{gurobi} with $M=10^5$. The simulation video is accessible via \href{https://youtu.be/a8gQCNuoNbM}{https://youtu.be/a8gQCNuoNbM}.
}{%
{We use Python 3.10.12 on a computer with 3.5 GHz Apple M2 Pro and 16G RAM. The Big-M based MILP is {\color{black}solved} using the Gurobi solver~\cite{gurobi} with $M=10^5$. The attached multimedia file contains the simulation video.}
}

 \subsection{Navigation Task}
The LTL specifications for task 2 are:
\begin{sizeddisplay}{\small}
\begin{align}\label{eq:task2}
  L_1: \quad & \level{1}{1} = \Diamond \left(\level{1}{2}  \wedge \Diamond \left(\level{2}{2} \wedge \Diamond \left(\level{3}{2}  \wedge  \Diamond  \level{4}{2} \right)\right)\right) \nonumber \\
  L_2: \quad & \level{1}{2} = \Diamond \pi_{\text{furn}}^{1,1} \wedge \Diamond \pi_{\text{outd}}^{1,1}  \nonumber \\
     & \level{2}{2}  = \Diamond \pi_{\text{heal}}^{1,1} \wedge \Diamond \pi_{\text{groc}}^{1,1}  \\
    & \level{3}{2}  = \Diamond \pi_{\text{elec}}^{1,1} \wedge \Diamond \pi_{\text{pet}}^{1,1} \nonumber \\
    & \level{4}{2}  = \Diamond (\pi_{\text{pack}}^{1,1} \wedge \Diamond \pi_{\text{dock}}^{1,1}) \nonumber 
\end{align}
\end{sizeddisplay}
These require that 1) A type 1 robot should {\it initiate} its task by gathering items from the furniture and outdoor sections {\it in any order}; 2) {\it Following} that the robot gathers items from the health and grocery sections {\it in any order}; 3) {\it Subsequently} the robot gathers items from the electronics and pet sections again {\it without any specific sequence}; 4) {\it After} delivering all the items to the packing area, the robot {\it eventually} returns to the dock.
The LTL specifications for task 3 are:
\begin{sizeddisplay}{\small}
\begin{align}\label{eq:task3}
 L_1: \quad & \level{1}{1} = \Diamond (\level{1}{2} \wedge \Diamond \level{2}{2}) \wedge (\Diamond \level{3}{2}  \vee \Diamond \level{4}{2} ) \nonumber \\
  L_2:  \quad   & \level{1}{2}  = \Diamond \pi_{\text{heal}}^{1,1} \wedge \Diamond \pi_{\text{groc}}^{1,1} \wedge \Diamond \pi_{\text{elec}}^{1,1}  \wedge \Diamond \pi_{\text{pet}}^{1,1}  \nonumber  \\
   & \level{2}{2}  = \Diamond (\pi_{\text{pack}}^{1,1} \wedge \Diamond \pi_{\text{dock}}^{1,1}) \\
  &  \level{3}{2} = \Diamond (\pi_{\text{outd}}^{2,2} \wedge \Diamond (\pi_{\text{pack}}^{2,2} \wedge \Diamond \pi_{\text{dock}}^{2,2}))\nonumber \\
    & \level{4}{2}  =  \Diamond (\pi_{\text{outd}}^{3,3} \wedge \Diamond (\pi_{\text{pack}}^{3,3} \wedge \Diamond \pi_{\text{dock}}^{3,3}))\nonumber 
\end{align}
\end{sizeddisplay}
These require that 1) A type 1 robot gathers items from the health, grocery, electronics and pet sections {\it in any order}; 2) {\it Either} a type 2 robot {\it or} a type 3 robot gathers items from the outdoor section; 3) {\it After} the items are delivered to the packing area, all robots {\it eventually} return to the dock.

\ifthenelse{\boolean{arXiv}}%
{%
 \begin{figure}[!t]
    \centering
        \subfigure[Task network of task 2]{
      \label{fig:task2}
      \includegraphics[width=\linewidth]{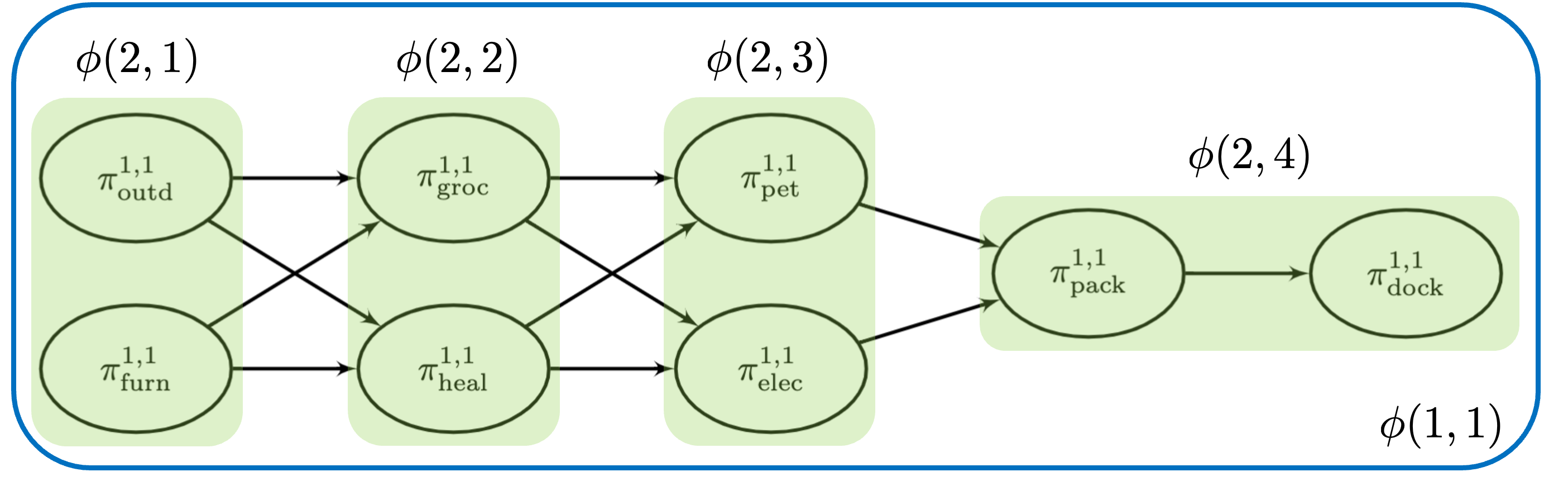}}
    \subfigure[Task network of task 3]{
      \label{fig:task3}
      \includegraphics[width=\linewidth]{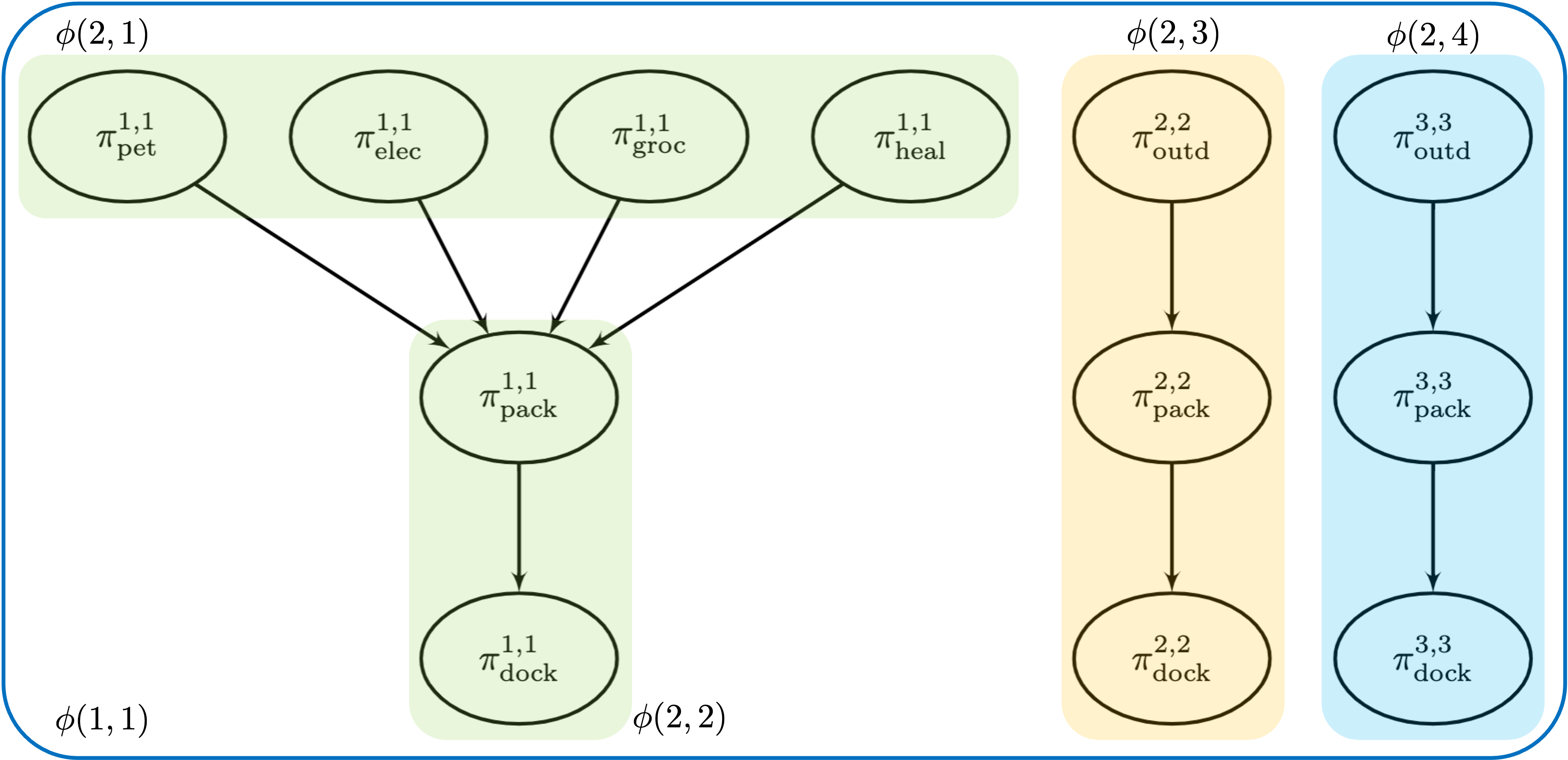}}
    \caption{Task networks. All nodes have self-loops labelled with $\top$ and labels are omitted for the sake of simplicity.}
    \label{fig:tasks23}
 \end{figure}
The task networks are shown in Fig.~\ref{fig:tasks23}. Flat LTL specifications can be found in Eqs.~\eqref{eq:flat_task2} and~\eqref{eq:flat_task3} in Appendix~\ref{app:task}, respectively.
}{%
The task networks and flat versions of these two tasks are in~\cite{luo2023robotic}.
}
We use~\cite{luo2022temporal} to tackle the flat specifications. As far as we know, the only method for hierarchical LTL is in~\cite{luo2024simultaneous}, but it's unsuitable here. First,~\cite{luo2024simultaneous} assumes robots work independently within a flat specification, which forbids robot coordination, like in $\level{1}{3}$ in Eq.~\eqref{eq:mrpd}. Second, that method requires that sequential sub-tasks in a flat specification must be completed by the same robot. Finally, it cannot address specifications that explicitly require sub-tasks to be performed by the same robot, as in $\level{1}{2}$ in Eq.~\eqref{eq:task3}.

\renewcommand{\arraystretch}{1.2}
\begin{table}[!t]
\centering
\begin{tabular}{|c||c|c||c|c|}
\hline
task &  $l_{\text{flat}}$ &  $l_{\text{hier}}$ &  $\ccalA_{\text{flat}}$ &  $\ccalA_{\text{hier}}$\\ 
\hline
1 &  51 & {\bf 35} & (387, 13862) &  {\bf (33, 45)} \\ 
\hline
2 & 45 & {\bf 19} & (12, 63) &  {\bf (20, 28)}  \\ 
\hline
3 & 35 & {\bf 27} & (113, 1891) &  {\bf (35, 101)}   \\ 
\hline\hline 
task & $t_{\text{flat}}$ &  $t_{\text{hier}}$ & $c_{\text{flat}}$ &  $c_{\text{hier}}$\\
\hline
1 &  126.1$\pm$2.7 & {\bf 18.6$\pm$0.8} & 289.3$\pm$3.0 & {\bf 237.9$\pm$3.5}\\
\hline
2 &  30.8$\pm$0.5 &  {\bf 23.8$\pm$1.3} & 115.6$\pm$1.9 & {\bf 114.9$\pm$1.6}\\
\hline
3 &  25.8$\pm$1.8  & {\bf 20.1$\pm$0.8} & 148.1$\pm$2.1 & {\bf 147.9$\pm$1.9}\\
\hline
\end{tabular}
\caption{The length of LTL specifications is represented by $l_{\text{flat}}$ and $l_{\text{hier}}$, while the size of the NBA is denoted by $\ccalA_{\text{flat}}$ and $\ccalA_{\text{hier}}$, specifying the number of nodes first, followed by the number of edges. The runtimes are indicated by $t_{\text{flat}}$ and $t_{\text{hier}}$, and the plan horizons are indicated by $c_{\text{flat}}$ and 
$c_{\text{hier}}$.}
\label{tab:result}
\end{table}

The {\it length} of a LTL specification is quantified by the number of operators it contains, including both logical and temporal operators~\cite{baier2008principles}. For each task, we compare between the length of the LTL and the size of the NBA, taking into account both the flat and hierarchical forms. These NBAs are built using LTL2BA~\cite{gastin2001fast}. In the case of the hierarchical form, the total is the sum of all the specifications. The initial locations for the robots are randomly sampled within the dock, and the runtimes and plan horizons are averaged over 20 runs. The statistical outcomes are displayed in Tab.~\ref{tab:result}. The hierarchical form of LTL specifications leads to more concise formulas in length and NBAs that are reduced in size with fewer nodes and edges. The reason is that in these three tasks, composite sub-tasks, which can encompass several atomic sub-tasks, have precedence relations between them. While expressing these relationships using flat specifications can be wordy due to the need for pairwise enumeration, using hierarchical LTL allows for a much more succinct representation.
The reduction in complexity leads to decreased runtimes and costs, particularly for task 1, where it notably minimizes both the runtimes and horizons.

 \subsection{Manipulation Task}

{\color{black}Two manipulation arms of the same type are tasked to assemble two LEGO models in the Gazebo simulator. The task specifications are hard to specify  by flat specifications.} To create executable trajectories, we use an online planning approach. Upon the completion of a sub-task, thereby freeing an arm, we evaluate the subsequent sub-task in the plan allocated to be executed by this arm. If this sub-task is not preceded by any other ongoing or pending sub-tasks, it will be carried out by this arm. However, if there are preceding sub-tasks that are yet to be completed, the sub-task is deferred until all preceding tasks are finished.
\ifthenelse{\boolean{arXiv}}%
{%
 \begin{figure}[!t]
    \centering
        \subfigure[Bin packing]{
      \label{fig:iclltl}
      \includegraphics[width=0.45\linewidth]{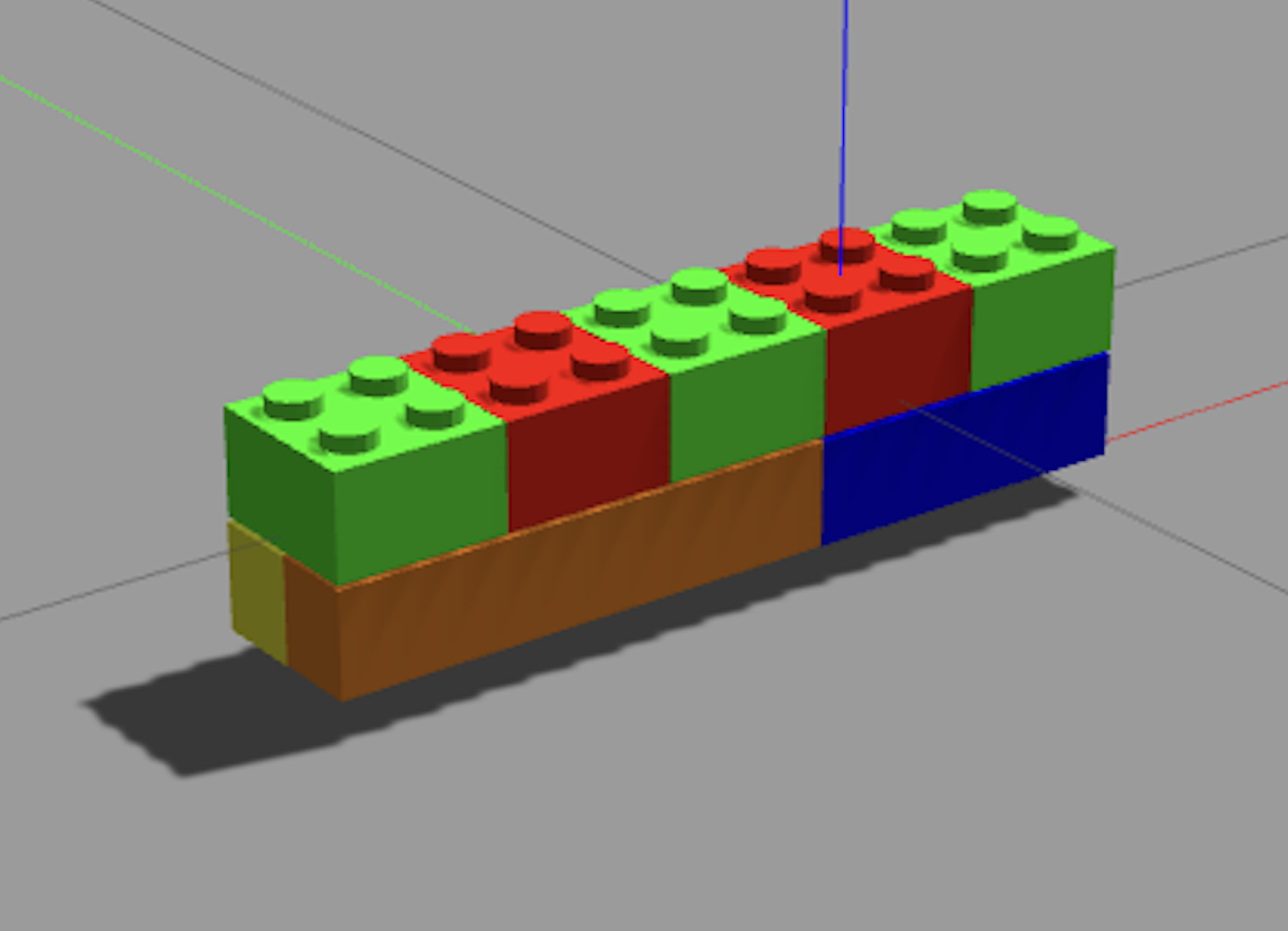}}
    \subfigure[LEGO model house]{
      \label{fig:house}
      \includegraphics[width=0.45\linewidth]{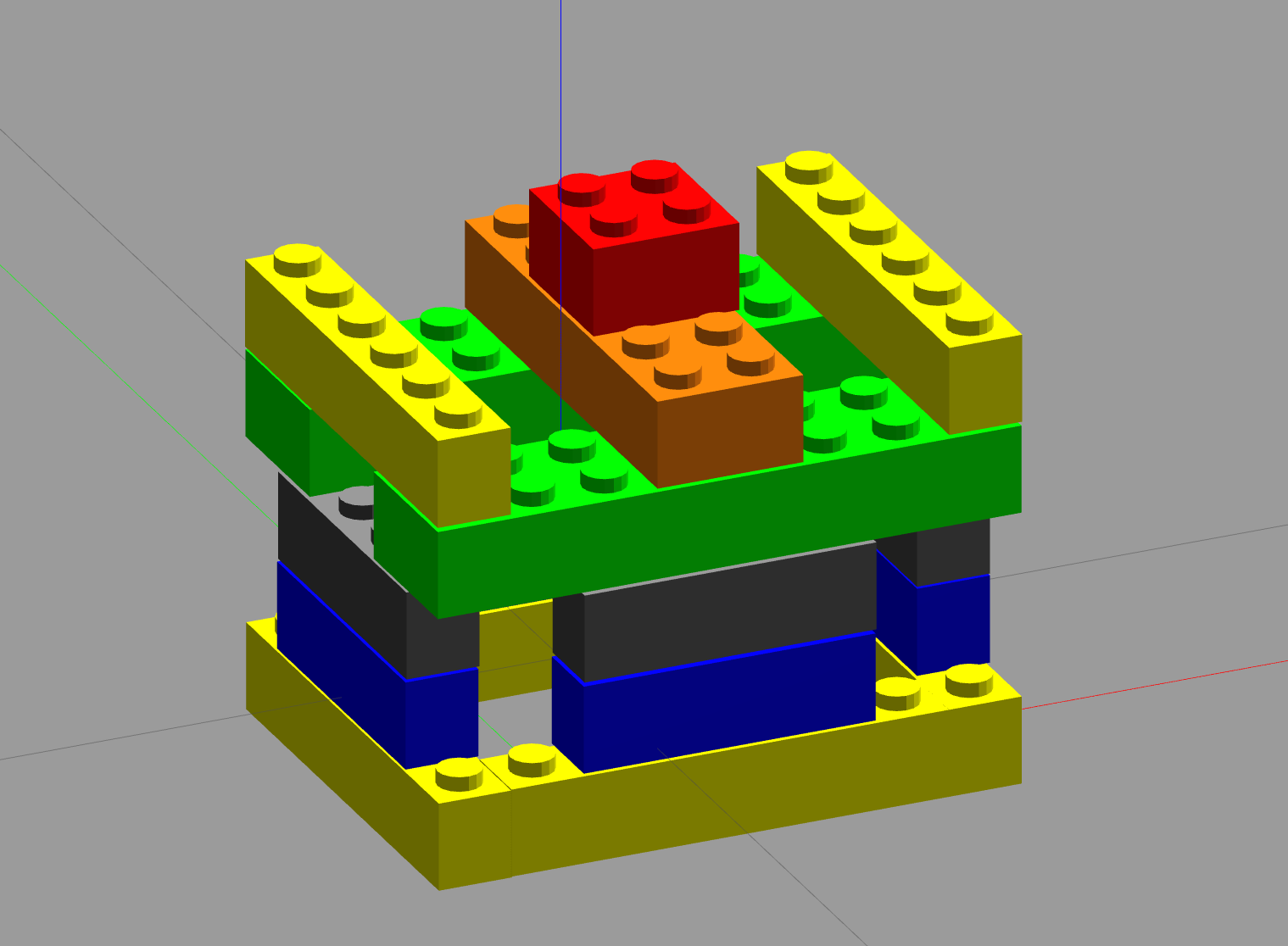}}
    \caption{Two LEGO models.}
    \label{fig:arm_model}
 \end{figure}
 }{
 \begin{figure}[!t]
    \centering
        \subfigure[Bin packing]{
      \label{fig:iclltl}
      \includegraphics[width=0.4\linewidth]{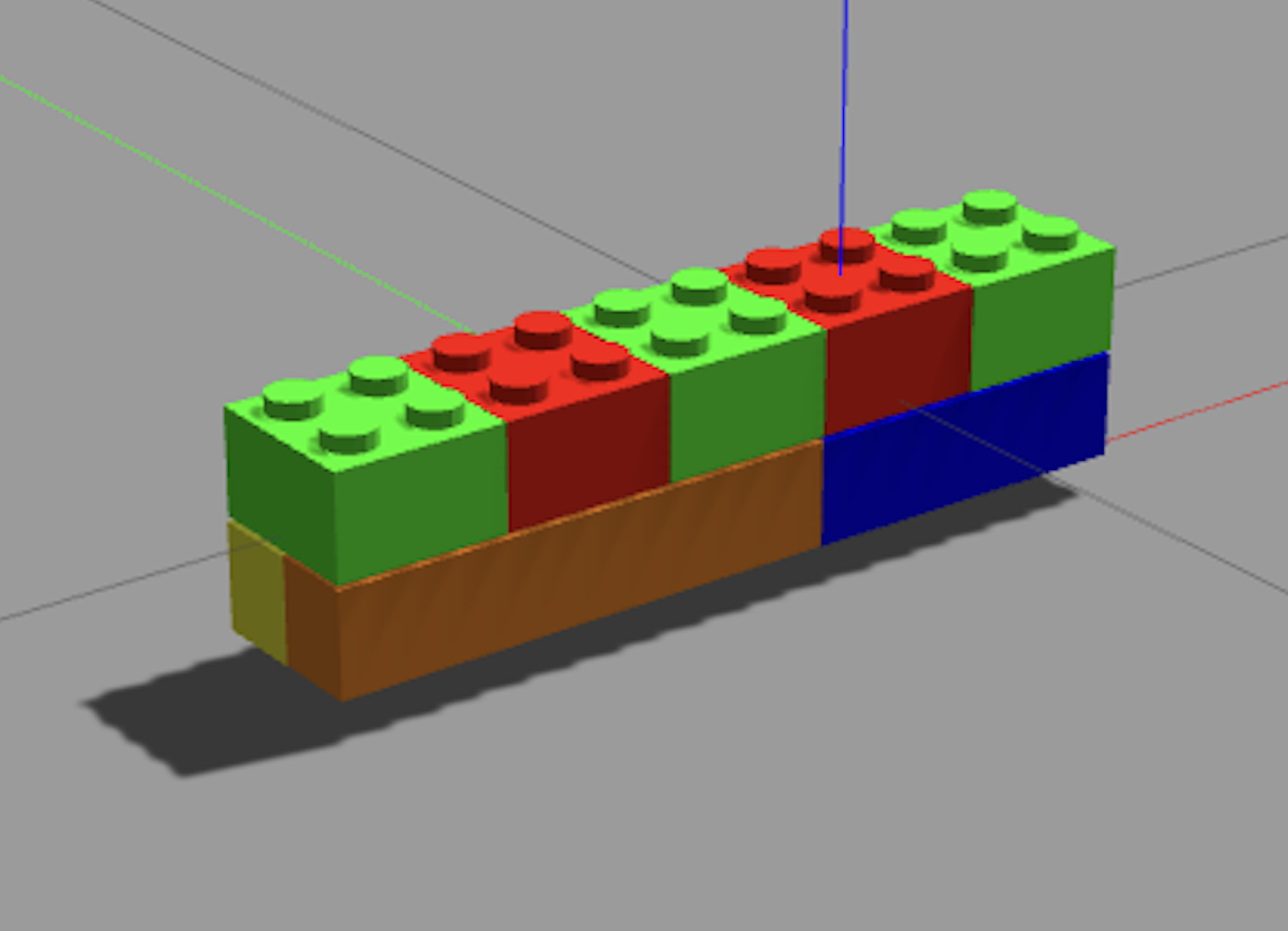}}
      \hspace{0.5cm}
    \subfigure[LEGO model house]{
      \label{fig:house}
      \includegraphics[width=0.4\linewidth]{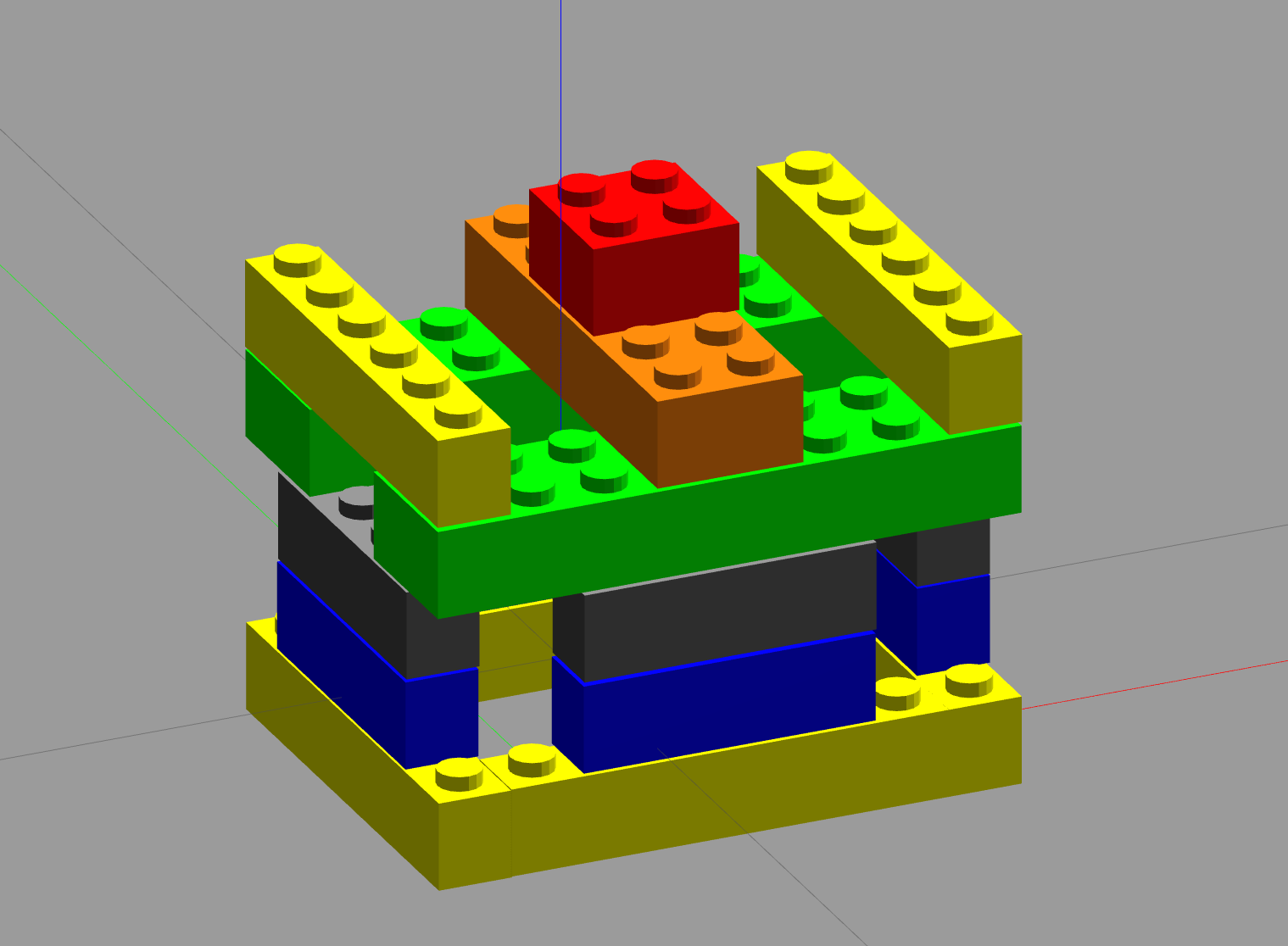}}
    \caption{Two LEGO models.}
    \label{fig:arm_model}
 \end{figure}
 }
\subsubsection{Bin packing} We use the LEGO model to replicate a bin packing task, as depicted in Fig.~\ref{fig:iclltl}. This task involves two adjacent stacks of bins, with the requirement to pack two stacks {\it in any order}, ensuring that the second level is packed only {\it after} the first level for each stack:
\begin{sizeddisplay}{\small}
\begin{align}
    L_1: \quad & \level{1}{1} = \Diamond \level{1}{2} \wedge  \Diamond \level{2}{2}   \nonumber \\
L_2: \quad & \level{1}{2} =  \Diamond \level{1}{3} \wedge   \Diamond \level{2}{3} \wedge   \neg  \level{2}{3} \, \mathcal{U}\, \level{1}{3} \nonumber \\
 & \level{2}{2} =  \Diamond \level{3}{3} \wedge   \Diamond \level{4}{3} \wedge   \neg   \level{4}{3} \, \mathcal{U}\, \level{3}{3} \nonumber \\
L_3: \quad &\begin{rcases}
 \level{1}{3} =  \Diamond \pi_{\tdot{yellow}} \wedge  \Diamond \pi_{\tdot{orange}}  \quad \quad \quad  
\end{rcases}  \text{1st level of stack 1}\\
&\begin{rcases}
            \level{2}{3} =  \Diamond\pi_{\tdot{green}} \wedge  \Diamond\pi_{\tdot{red}}  \wedge   \Diamond \pi_{\tdot{green}}  \; \nonumber 
\end{rcases}  \text{2nd level of stack 1}\\
&\begin{rcases}
           \level{3}{3} =  \Diamond  \pi_{\tdot{blue}} \wedge \Diamond \pi_{\tdot{gray}}   \quad \quad \quad \nonumber  
\end{rcases}  \text{1st level of stack 2} \\
&\begin{rcases}
           \level{4}{3} =  \Diamond  \pi_{\tdot{red}} \wedge \Diamond \pi_{\tdot{green}}   \quad \quad \quad \nonumber  
\end{rcases}  \text{2nd level of stack 2}
\end{align}
\end{sizeddisplay}
where $\level{1}{2}$ and $\level{2}{2}$ represent packing two stacks of bins, while $\pi_{\tdot{blue}}$ denotes packing a bin colored $\tdot{blue}$ by any arm. Since both arms are of the same type, the atomic propositions do not carry a superscript to denote the robot type. There is a total of 30 states and 44 edges in the NBAs. It took \(4.9 \pm 0.2\) seconds over 20 runs to generate a task allocation plan.

\subsubsection{House model} The second task involves constructing a complex house model, level by level, as shown in Fig.~\ref{fig:house}. Since the model is symmetric when viewed from the front, each arm essentially handles half of the overall workload.
\ifthenelse{\boolean{arXiv}}%
{%
  The hierarchical LTL specifications are 
  \begin{sizeddisplay}{\small}
  \begin{align}
    L_1: \quad & \level{1}{1} = \Diamond (\level{1}{2} \wedge  \Diamond (\level{2}{2} \wedge \Diamond (\level{3}{2}  \nonumber \\
  & \quad \quad \quad \quad  \wedge \Diamond (\level{4}{2} \wedge \Diamond (\level{5}{2} \wedge \Diamond \level{6}{2} )))))\nonumber \\
L_2: \quad & \level{1}{2} = \Diamond \level{1}{3} \wedge \Diamond \level{2}{3} \nonumber \\
\quad & \level{2}{2} = \Diamond \level{3}{3} \wedge \Diamond \level{4}{3} \nonumber \\
\quad & \level{3}{2} = \Diamond \level{5}{3} \wedge \Diamond \level{6}{3} \nonumber \\
\quad & \level{4}{2} = \Diamond \level{7}{3} \nonumber \\
\quad & \level{5}{2} = \Diamond (\level{8}{3} \wedge \Diamond \level{9}{3}) \nonumber \\
\quad & \level{6}{2} = \Diamond \level{10}{3} \\
 L_3:  \quad 
& \begin{rcases}
  \level{1}{3}  = \Diamond \pi_{\tdot{yellow}}^{1}  \wedge \Diamond \pi_{\tdot{yellow}}^{1} \nonumber \\  
 \level{2}{3} = \Diamond \pi_{\tdot{yellow}}^{2}  \wedge \Diamond \pi_{\tdot{yellow}}^{2} \;\;\; \quad \nonumber \\
\end{rcases}  \text{1st level} \nonumber \\
& \begin{rcases}
 \level{3}{3} = \Diamond (\pi_{\tdot{blue}}^{1}  \wedge \Diamond \pi_{\tdot{gray}}^{1}) \nonumber \\
  \level{4}{3} = \Diamond (\pi_{\tdot{blue}}^{2}  \wedge \Diamond \pi_{\tdot{gray}}^{2}) \nonumber \\
  \level{5}{3} = \Diamond (\pi_{\tdot{blue}}^{1}  \wedge \Diamond \pi_{\tdot{gray}}^{1}) \nonumber \\
  \level{6}{3} = \Diamond (\pi_{\tdot{blue}}^{2}  \wedge \Diamond \pi_{\tdot{gray}}^{2}) \quad \nonumber \\
\end{rcases} \text{2nd and 3rd levels} \nonumber \\
& \begin{rcases}
\level{7}{3} = \Diamond \pi_{\tdot{green}}  \wedge \Diamond \pi_{\tdot{green}} \quad  \;\;\;\nonumber \\
\end{rcases} \text{4th level} \nonumber \\
& \begin{rcases}
 \level{8}{3} = \Diamond \pi_{\tdot{yellow}}^{1}  \wedge \Diamond \pi_{\tdot{yellow}}^{2} \quad\;\;\; \nonumber \\
 \level{9}{3} = \Diamond \pi_{\tdot{orange}} \nonumber \\
\end{rcases}  \text{5th level} \nonumber \\
& \begin{rcases}
\level{10}{3} = \Diamond \pi_{\tdot{red}}  \quad \quad \quad \quad\,
\end{rcases} \text{6th level} \nonumber 
\end{align}
  \end{sizeddisplay}
where the identical superscript represents that blocks are assembled by the same arm. For instance, leaf specifications $\level{1}{3}$ and $\level{2}{3}$ specify the assembly of the 1st level by assigning two yellow blocks to each arm, and $\level{10}{3}$ specify that the red block at the top can be assembled by any arm.
}{%
  The complete hierarchical LTL specifications are available in~\cite{luo2023robotic}.
}
{\color{black}There is a total of 58 states and 59 edges in all NBAs}. It took \(4.8 \pm 0.2\) seconds over 20 runs to generate a task allocation plan.


\section{Discussion and Conclusions}
In this work, we presented a hierarchical decomposition-based method to address robotic navigation and manipulation tasks as described by hierarchical LTL specifications. Although our proposed approach demonstrates effectiveness, it falls short in terms of completeness. Consequently, developing an approach that guarantees both completeness and optimality stands as a potential avenue for future research.


\begin{sizeddisplay}{\small}
\bibliographystyle{IEEEtranN}
\bibliography{IEEEabrv,xl_bib}
\end{sizeddisplay}
\appendices
{\color{black}
\section{} \label{app:milp}
\subsection{MILP objective}
Drawing inspiration from VRP, we construct a routing graph $(\ccalV_r, \ccalE_r)$ in which nodes denote regions (e.g., a grocery shelf, a brick target location) associated with certain atomic sub-task, and the traversal of nodes within some duration by appropriate robots fulfills certain atomic propositions. The goal is to minimize the $\alpha$-weighted sum of the energy cost and completion time:
\begin{align}
\min_{x_{uvr}, t_e} \; \alpha \sum_{(u,v)\in \ccalE_r} \sum_{r \in \ccalM(v)} d_{uv} x_{uvr} + (1 - \alpha) \sum_{e} t_{e}.
\end{align}
Here $x_{uvr} \in \{0, 1\}$ are routing variables indicating the movement of robot $r$ between nodes $u$ and $v$ in the routing graph, with robot $r$ being part of the subset $\ccalM(v)$ which assigns a type of robots with the required skills to a node, and $d_{uv}$ is the energy cost. Additionally, $e$ stands for the atomic sub-tasks within the task network.

We present the concepts of activation and completion times and identify three categories of temporal constraints. 

 \begin{defn}[Activation and completion time of a sub-task or its starting vertex label~\cite{luo2022temporal}]\label{defn:time}
 For a given sub-task \( e = (v_1, v_2) \), its activation time (or the activation time of its starting vertex label) is defined as the moment when the vertex label \( \gamma(v_1) \) become satisfied. Similarly, the completion time of a sub-task (or the completion time of its starting vertex label) is determined as the moment when its edge label \( \gamma(v_1, v_2) \) is satisfied (or alternatively, the final moment when its starting vertex label \( \gamma(v_1) \) remains true). The span of a sub-task (or its starting vertex label) is the time period that begins with the activation time and ends with the completion time.
 \end{defn}

Let \( t^-_{ep} \) and \( t^+_{ep} \) denote the first and last time instants at which the \( p \)-th clause in the starting vertex label of sub-task \( e \) is satisfied, respectively. Considering two atomic sub-tasks \( e \) and \( e' \), let \( \phi \) and \( \phi' \) be their corresponding leaf specifications. The linear inequality constraints \(\langle 13 \rangle\), \(\langle 17 \rangle\), \(\langle 18 \rangle\), \(\langle 19 \rangle\), \(\langle 23 \rangle\), and \(\langle 25 \rangle\) are modified with modifications being highlighted in boxes. Notations are simplified for ease of understanding.
 
\subsection{Temporal constraints associated with one atomic sub-task}
Constraint \(\langle 13 \rangle\) captures the relation that the completion time of a sub-task must fall within the duration of its starting vertex label, or precisely one time step subsequent to the completion of this starting vertex label, which becomes:
\begin{subequations}\label{eq:span}
     \begin{align}
  t_{ep}^-   & \leq  t_e + \modf{M(1 - b_{\phi})}\\ 
     t_e   \leq t_{ep}^+ + 1  + M & (1 - b_{p}) + \modf{M (1 - b_{\phi})}. \label{eq:17}
   \end{align}
\end{subequations}
Constraints are activated only if \( b_\phi = 1 \) and \( b_p = 1 \), then the condition \( t_{ep}^- \leq t_e \leq t_{ep}^+ + 1 \) come into effect.


\subsection{Temporal constraints associated with the completion of a sub-task and the activation of the sub-task immediately following it} 
Constraint \(\langle 17 \rangle\) reflecting the precedence rule that sub-task \( e \) must be completed at most one time step prior to the activation of its immediately subsequent sub-task, becomes
\begin{align}\label{eq:after}
   t_e + 1 \leq  t_{e'} & + M (1 - b_{ee'}) \nonumber \\ 
  & +  \modf{M (1 - b_\phi) + M (1 - b_{\phi'})},\; \forall\, e' \prec e,
\end{align}
where the binary variable \( b_{ee'} \) is set to 1 if sub-task \( e' \) directly follows sub-task \( e \). Furthermore, constraint \eqref{eq:after} is activated only when both \( b_\phi \) and \( b_{\phi'} \) equal 1, leading to \( t_e + 1 \leq t_{e'} \).

Constraint \(\langle 18 \rangle\) states that the sub-task \( e' \), if following right after sub-task \( e \), must be activated no more than one time step after the completion of \( e \),  which becomes
 \begin{align}\label{eq:20}
   t_{e'p}^-  \leq t_{e} & + 1 + M (1 - b_{ee'}) \nonumber \\ 
  & +  \modf{M (1 - b_\phi) + M (1 - b_{\phi'})}.
 \end{align}
When \( b_{ee'} = 1 \), indicating that sub-task \( e' \) immediately follows sub-task \( e \), and additionally, when \( b_\phi = b_{\phi'} = 1 \), the condition \( t_{e'p}^- \leq t_{e} + 1 \) is established.

Constraint \(\langle 19 \rangle\) implies that two sub-tasks cannot be completed simultaneously, which is now expressed as:
\begin{subequations}
    \begin{align}
             b_e^{e'} + b_{e'}^e &  = 1, \label{eq:diff_a}\\
        M (b_{e}^{e'} - 1) & \leq t_e -  t_{e'} + \modf{M (1 - b_\phi) + M (1 - b_{\phi'})}, \\ 
       t_e -  t_{e'}   \leq & M b_{e}^{e'} - 1 + \modf{M (1 - b_\phi) + M (1 - b_{\phi'})}, \label{eq:diff_b}
    \end{align}
\end{subequations}
where \( b_{e}^{e'} = 1 \) indicates that sub-task \( e \) is completed after sub-task \( e' \). When \( b_\phi = b_{\phi'} = 1 \), and if \( b_{e}^{e'} = 1 \), it follows that \( t_e \geq t_{e'} \). Meanwhile, if \( b_{e'}^{e} = 0 \), the condition \( t_{e'} - t_e \leq -1 \) applies. Consequently, this leads to  \( t_e \geq t_{e'} + 1 \).

\subsection{Temporal constraints associated with the activation of the first sub-task} Constraint \(\langle 23 \rangle\) specifies that if \( e \) is the first one to be completed and there is no preceding sub-task whose completion triggers \( e \), then \( e \) must be activated at the beginning:
 \begin{align}\label{eq:zeroactivation}
   t_{ep}^-  \leq   M (1 - b_e) + M (1- b_p) + \modf{ M (1 - b_{\phi})},
 \end{align}
where the binary variable \( b_e = 1 \) indicates that sub-task \( e \) is the first to be completed. When \( b_\phi = b_e = b_p = 1 \), it follows that \( t_{ep}^- = 0 \), meaning that \( e \) is activated at the beginning with the \( p \)-th clause in the vertex label being true.

Constraint \(\langle 25 \rangle\) specifies the specific categories of states from which robots should initiate, which becomes:
\begin{subequations}\label{eq:routingforactivation}
   \begin{align}
     b_{pq}^{\text{init}} & \leq b_p, \\ 
     b_{pq}^{\text{init}} & \leq M  b_e + \modf{ M (1 - b_\phi)}, \label{eq:routingforactivation_b} \\
     b_{pq}^{\text{prior}} & \leq b_p,  \\
   b_{pq}^{\text{prior}} &\leq  M (1 - b_e)  +  \modf{ M (1 - b_\phi)}.  \label{eq:routingforactivation_a}
   \end{align}
 \end{subequations}
For the robots engaged in the \( q \)-th positive literal of the \( p \)-th clause in the starting vertex label of sub-task \( e \), the binary variable \( b_{pq}^{\text{init}} \) indicates that the robots begin from their initial states. Conversely, \( b_{pq}^{\text{prior}} \) indicates that the robots start from their positions following the completion of preceding sub-tasks of \( e \). If \( b_p = 0 \), it implies no robot is involved in the \( p \)-th clause, leading to \( b_{pq}^{\text{init}} = b_{pq}^{\text{prior}} = 0 \). When \( b_\phi = b_e = 1 \) and \( b_p = 1 \), it results in \( b_{pq}^{\text{prior}} = 0 \). This means that if sub-task \( e \) is the first to be completed, the involved robots should begin from their initial states. Conversely, when \( b_\phi = 1 \), \( b_e = 0 \), and \( b_p = 1 \), it leads to \( b_{pq}^{\text{init}} = 0 \), indicating that if sub-task \( e \) is not the first to be completed, the involved robots should start from their locations after finishing prior sub-tasks of \( e \).

\ifthenelse{\boolean{arXiv}}%
{%
\section{Flat specifications for MRPD tasks}\label{app:task}
\subsubsection{Task 1}
\begin{align}\label{eq:flat_task1}
       \phi = & \, \Diamond (\apsm{furn}{1}{1} \wedge \bigcirc(\apsm{furn}{1}{1} \, \mathcal{U}\,\apsm{furn}{3}{3}))  \nonumber\\
      & \wedge \Diamond(\apsm{pack}{3}{3} \wedge \Diamond \apsm{dock}{3}{3}) \wedge \neg \apsm{pack}{3}{3} \,\mathcal{U}\, \apsm{furn}{3}{3}  \nonumber \\
      & \wedge \Diamond \apsm{outd}{1}{1} \wedge \Diamond \apsm{pet}{1}{1} \wedge \Diamond (\apsm{pack}{1}{1} \wedge \Diamond \apsm{dock}{1}{1}) \nonumber \\
      & \wedge \neg \apsm{outd}{1}{1} \,\mathcal{U}\, \apsm{furn}{1}{1} \wedge \neg \apsm{outd}{1}{1} \,\mathcal{U}\, \apsm{furn}{3}{3}  \\
      & \wedge \neg \apsm{pet}{1}{1} \,\mathcal{U}\, \apsm{furn}{1}{1} \wedge \neg \apsm{pet}{1}{1} \,\mathcal{U}\, \apsm{furn}{3}{3}  \nonumber \\
      & \wedge \neg \apsm{pack}{1}{1} \,\mathcal{U}\, \apsm{outd}{1}{1} \wedge \neg \apsm{pack}{1}{1} \,\mathcal{U}\, \apsm{pet}{1}{1}  \nonumber \\
      & \wedge \Diamond \apsm{heal}{2}{2} \wedge \Diamond \apsm{groc}{2}{2} \wedge  \Diamond ( \apsm{pack}{2}{2} \wedge \Diamond \apsm{dock}{2}{2} )\nonumber \\
     & \wedge \neg \apsm{groc}{2}{2} \,\mathcal{U}\,  \apsm{heal}{2}{2} \wedge \neg \apsm{pack}{2}{2} \,\mathcal{U}\,  \apsm{groc}{2}{2} \nonumber 
\end{align}
\subsubsection{Task 2}
\begin{align}\label{eq:flat_task2}
      \phi =  &\, \Diamond \apsm{furn}{1}{1} \wedge \Diamond \apsm{outd}{1}{1} \wedge \Diamond \apsm{heal}{1}{1} \wedge \Diamond \apsm{groc}{1}{1} \nonumber \\ 
      & \wedge\Diamond \apsm{elec}{1}{1}  \wedge \Diamond \apsm{pet}{1}{1}   \wedge \Diamond (\apsm{pack}{1}{1}   \wedge \Diamond \apsm{dock}{1}{1}) \nonumber \\
      & \wedge \neg \apsm{heal}{1}{1} \,\mathcal{U}\,  \apsm{furn}{1}{1} \wedge \neg \apsm{heal}{1}{1} \,\mathcal{U}\,  \apsm{outd}{1}{1} \nonumber \\
      & \wedge \neg \apsm{groc}{1}{1} \,\mathcal{U}\,  \apsm{furn}{1}{1} \wedge \neg \apsm{groc}{1}{1} \,\mathcal{U}\,  \apsm{outd}{1}{1}  \\
      & \wedge \neg \apsm{elec}{1}{1} \,\mathcal{U}\,  \apsm{heal}{1}{1} \wedge \neg \apsm{elec}{1}{1} \,\mathcal{U}\,  \apsm{groc}{1}{1}  \nonumber \\
      & \wedge \neg \apsm{pet}{1}{1} \,\mathcal{U}\,  \apsm{heal}{1}{1} \wedge \neg \apsm{pet}{1}{1} \,\mathcal{U}\,  \apsm{groc}{1}{1} \nonumber \\
      & \wedge \neg \apsm{pack}{1}{1} \,\mathcal{U}\,  \apsm{elec}{1}{1} \wedge \neg \apsm{pack}{1}{1} \,\mathcal{U}\,  \apsm{pet}{1}{1} \nonumber
\end{align}

\subsubsection{Task 3}
\begin{align}\label{eq:flat_task3}
    \phi = &\, \Diamond \pi_{\text{heal}}^{1,1} \wedge \Diamond \pi_{\text{groc}}^{1,1} \wedge \Diamond \pi_{\text{elec}}^{1,1}  \wedge \Diamond \pi_{\text{pet}}^{1,1}  \nonumber  \\
    & \wedge \Diamond (\pi_{\text{pack}}^{1,1} \wedge \Diamond \pi_{\text{dock}}^{1,1})\nonumber \\
    & \wedge  \neg \apsm{pack}{1}{1} \,\mathcal{U}\,  \apsm{heal}{1}{1} \wedge \neg \apsm{pack}{1}{1} \,\mathcal{U}\,  \apsm{groc}{1}{1}   \\
    & \wedge  \neg \apsm{pack}{1}{1} \,\mathcal{U}\,  \apsm{elec}{1}{1} \wedge \neg \apsm{pack}{1}{1} \,\mathcal{U}\,  \apsm{pet}{1}{1}  \nonumber \\
    & \wedge \left(\Diamond  \left(\pi_{\text{outd}}^{2,2} \wedge \Diamond  \left(\pi_{\text{pack}}^{2,2} \wedge \Diamond \pi_{\text{dock}}^{2,2}\right)\right) \right. \nonumber \\ 
    & \vee \left.\Diamond \left(\pi_{\text{outd}}^{3,3} \wedge \Diamond \left(\pi_{\text{pack}}^{3,3} \wedge \Diamond \pi_{\text{dock}}^{3,3}\right)\right)\right) \nonumber
\end{align}
}{

}

}
\end{document}